\documentclass[lettersize,journal]{IEEEtran}
\usepackage{amsmath,amsfonts}
\usepackage{algorithmic}
\usepackage{algorithm}
\usepackage{array}
\usepackage[caption=false,font=normalsize,labelfont=sf,textfont=sf]{subfig}
\usepackage{textcomp}
\usepackage{url}
\usepackage{verbatim}
\usepackage{graphicx}
\usepackage{cite}
\usepackage{color}
\usepackage{diagbox}
\usepackage{graphics}
\usepackage{epsfig}
\usepackage{threeparttable}
\usepackage{xspace}
\usepackage{siunitx}
\usepackage{graphicx}
\usepackage{amssymb}
\usepackage{color}
\usepackage[normalem]{ulem}
\usepackage{arydshln} 
\usepackage{multirow}
\usepackage{float}
\usepackage{amsfonts}
\usepackage{bm}
\usepackage[table]{xcolor}
\usepackage{colortbl}
\usepackage{diagbox}
\usepackage{rotating}
\usepackage{booktabs}
\usepackage{overpic}
\usepackage{enumitem}
\usepackage{colortbl}
\usepackage{dblfloatfix}
\usepackage[export]{adjustbox}
\usepackage{pifont}
\usepackage[pagebackref=false,breaklinks=false,colorlinks=true, bookmarks=false]{hyperref}
\usepackage{cleveref}
\usepackage{tikz}
\usepackage[edges]{forest}
\usepackage{caption} 
\usepackage{utfsym}

\newcommand*{\belowrulesepcolor}[1]{%
  \noalign{%
    \kern-\belowrulesep 
    \begingroup 
      \color{#1}%
      \hrule height\belowrulesep 
    \endgroup 
    \vspace{-0.03mm}
  }%
} 
\newcommand*{\aboverulesepcolor}[1]{%
  \noalign{%
  \vspace{-0.03mm}
    \begingroup 
      \color{#1}%
      \hrule height\aboverulesep 
    \endgroup 
    \kern-\aboverulesep 
  }%
}

\usepackage{listings}
\usepackage{etoolbox}
\makeatletter
\AfterEndEnvironment{algorithm}{\let\@algcomment\relax}
\AtEndEnvironment{algorithm}{\kern2pt\hrule\relax\vskip3pt\@algcomment}
\let\@algcomment\relax
\newcommand\algcomment[1]{\def\@algcomment{\footnotesize#1}}
\renewcommand\fs@ruled{\def\@fs@cfont{\bfseries}\let\@fs@capt\floatc@ruled
  \def\@fs@pre{\hrule height.8pt depth0pt \kern2pt}%
  \def\@fs@post{}%
  \def\@fs@mid{\kern2pt\hrule\kern2pt}%
  \let\@fs@iftopcapt\iftrue}
\makeatother

\usepackage{tabulary}
\newcolumntype{I}{!{\vrule width 1pt}}
\newcolumntype{x}[1]{>{\centering\arraybackslash}p{#1pt}}
\newcolumntype{y}[1]{>{\raggedright\arraybackslash}p{#1pt}}
\newcolumntype{z}[1]{>{\raggedleft\arraybackslash}p{#1pt}}

\definecolor{mygray}{gray}{.85}
\definecolor{mygray1}{gray}{.7}
\definecolor{mygray2}{gray}{.93}
\definecolor{codegreen}{RGB}{79,126,127}
\definecolor{codedefine}{RGB}{153,54,159}
\definecolor{codefunc}{RGB}{73,122,234}
\definecolor{codecall}{RGB}{73,122,234}
\definecolor{codepro}{RGB}{212,96,80}
\definecolor{codedim}{RGB}{89,152,195}
\definecolor{3dgc1}{RGB}{177, 83, 74}
\definecolor{3dgc2}{RGB}{93, 107, 72}
\definecolor{hidden-draw}{RGB}{202,203,207}
\definecolor{hidden-2d}{RGB}{243,250,244}
\definecolor{hidden-video}{RGB}{250,248,243}
\definecolor{hidden-3d}{RGB}{236,249,253}
\definecolor{hidden-4d}{RGB}{244,243,250}

\makeatletter
\newcommand{\thickhline}{%
    \noalign {\ifnum 0=`}\fi \hrule height 1pt
    \futurelet \reserved@a \@xhline
}
\makeatother

\makeatletter
\DeclareRobustCommand\onedot{\futurelet\@let@token\@onedot}
\def\@onedot{\ifx\@let@token.\else.\null\fi\xspace}

\makeatother

\newcommand{\myPara}[1]{\vspace{.05in}\noindent\textbf{#1}}

\usepackage{etoolbox}
\usepackage{xcolor}
\usepackage{soul}
\usepackage{ifthen}
\newboolean{showcomments}
\setboolean{showcomments}{true}

\begin{document}
\title{Simulating the Real World: A Unified Survey of Multimodal Generative Models}

\author{Yuqi Hu$^\ast$, Longguang Wang$^\ast$, Xian Liu$^\ast$, Ling-Hao Chen$^\ast$, Yuwei Guo$^\ast$, 
Yukai Shi$^\ast$, Ce Liu$^\ast$, \\Anyi Rao, Zeyu Wang, and Hui Xiong$^\dag$,~\IEEEmembership{Fellow,~IEEE}

\vspace{0.3cm}
\textit{(Survey Paper)}

\thanks{Yuqi Hu, Zeyu Wang, and Hui Xiong are with the Thrust of Artificial Intelligence, The Hong Kong University of Science and Technology (Guangzhou), Guangzhou, China. Hui Xiong is also with the Department of Computer Science and Engineering, The Hong Kong University of Science and Technology Hong Kong SAR, China.
(e-mail: yhu873@connect.hkust-gz.edu.cn; zeyuwang@ust.hk; xionghui@ust.hk).}
\thanks{Anyi Rao is with the MMLab, The Hong Kong University of Science and Technology, Hong Kong, China.
(e-mail: anyirao@ust.hk).}
\thanks{Longguang Wang is with the School of Electronics and Communication Engineering, Shenzhen Campus of Sun Yat-sen University, Sun Yat-sen University, Shenzhen, China.
(e-mail: wanglg9@mail.sysu.edu.cn).}
\thanks{Xian Liu and Yuwei Guo are with The Chinese University of Hong Kong, Hong Kong, China.
(e-mail: alvinliu@ie.cuhk.edu.hk; guoyw@ie.cuhk.edu.hk).}
\thanks{Ling-Hao Chen and Yukai Shi are with Tsinghua University, Guangdong, China. 
(e-mail: evan@lhchen.top; shiyk22@mails.tsinghua.edu.cn).}
\thanks{Ce Liu is with Bosch (China) Investment Co., Ltd., Shanghai, China. 
(e-mail: celiu0901@gmail.com).}

\thanks{A project associated with this survey is available at \url{https://github.com/ALEEEHU/World-Simulator}.}

\thanks{
$\ast$ Equal contribution. 
$^\dag$ Corresponding author.
}

}

\markboth{Journal of \LaTeX\ Class Files,~Vol.~14, No.~8, August~2021}%
{Shell \MakeLowercase{\textit{et al.}}: A Sample Article Using IEEEtran.cls for IEEE Journals}

\maketitle
\begin{abstract} 
Understanding and replicating the real world is a critical challenge in Artificial General Intelligence (AGI) research. To achieve this, many existing approaches, such as world models, aim to capture the fundamental principles governing the physical world, enabling more accurate simulations and meaningful interactions. However, current methods often treat different modalities, including 2D (images), videos, 3D, and 4D representations, as independent domains, overlooking their interdependencies. Additionally, these methods typically focus on isolated dimensions of reality without systematically integrating their connections. In this survey, we present a unified survey for multimodal generative models that investigate the progression of data dimensionality in real-world simulation. Specifically, this survey starts from 2D generation (\textcolor[rgb]{0.5,0.8,0.7}{appearance}), then moves to video (\textcolor[rgb]{0.5,0.8,0.7}{appearance}+\textcolor{orange}{dynamics}) and 3D generation (\textcolor[rgb]{0.5,0.8,0.7}{appearance}+
\textcolor[rgb]{0.25, 0.5, 0.75}{geometry}), and finally culminates in 4D generation that integrate all dimensions. To the best of our knowledge, this is the first attempt to systematically unify the study of 2D, video, 3D, and 4D generation within a single framework. To guide future research, we provide a comprehensive review of datasets, evaluation metrics, and future directions to foster insights for newcomers. This survey serves as a bridge to advance the study of multimodal generative models and real-world simulation within a unified framework.

\end{abstract}
\begin{IEEEkeywords}
Generative models, image generation, video generation, 3D generation, 4D generation.
\end{IEEEkeywords}

\vspace{1cm}

\IEEEraisesectionheading{\section{Introduction}\label{sec:intro}}

\begin{figure*}[!t]
  \centering
  \includegraphics[width=1.0\linewidth]{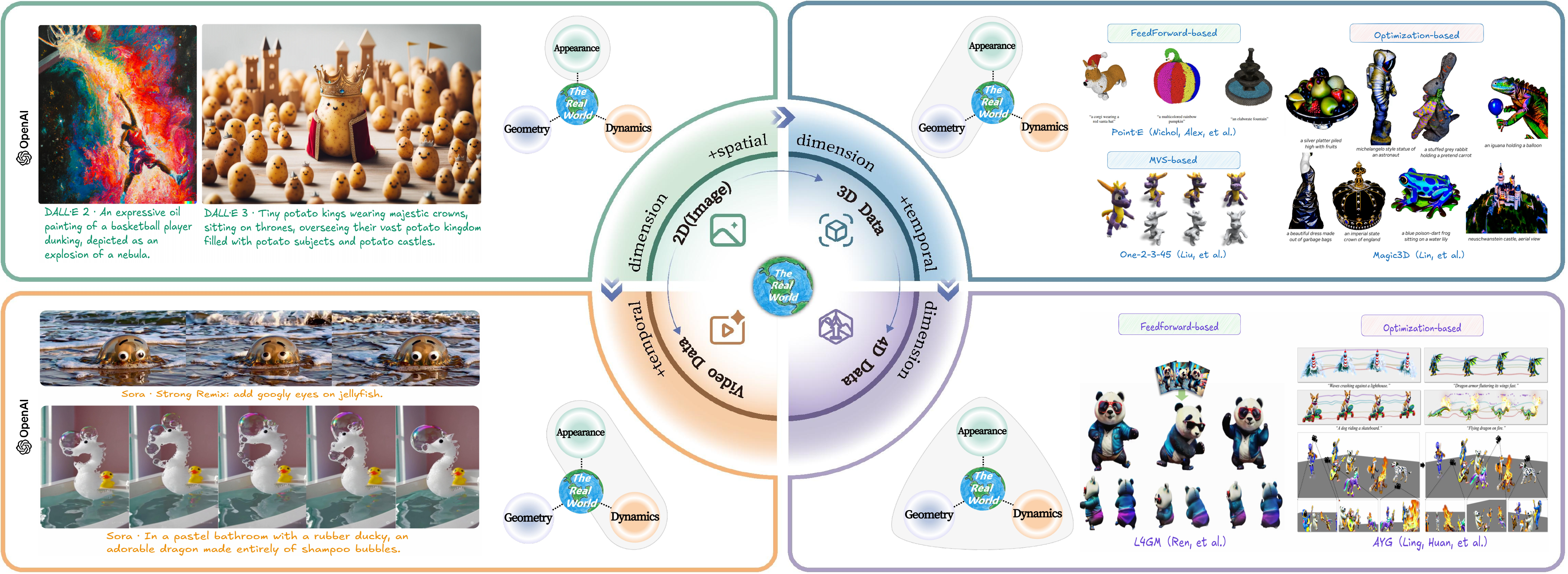}
  \caption{Roadmap of dimensional growth from 2D images to video, 3D, and 4D content in real-world simulation, outlining a conceptual taxonomy based on the coverage of data properties (i.e., appearance, geometry, and dynamics).}
  \label{fig:pipeline}
\end{figure*}

\IEEEPARstart{F}OR decades, the research community has aspired to develop systems that encapsulate the fundamental principles of the physical world, a cornerstone in the journey towards Artificial General Intelligence (AGI)~\cite{lecun2022path}. Central to this endeavor is simulating the real world with machines, aiming to capture the complexities of reality through multimodal generative models. The resultant world simulator holds the promise of advancing the understanding of the real world, unlocking transformative applications such as virtual reality~\cite{lee2024all}, games~\cite{bruce2024genie}, robotics~\cite{wu2023daydreamer}, and autonomous driving~\cite{wang2024driving}.

The term ``world simulator'' was first introduced by Ha David~\cite{ha2018world}, drawing an analogy to the concept of a mental model~\cite{forrester1971counterintuitive} in cognitive science. Building on this perspective, modern researchers formulate the simulator as an abstract framework that enables intelligent systems to simulate the real world through multimodal generative models. These models encode visual contents and spatial-temporal dynamics of the real world into compact representations. As geometry, appearance, and dynamics jointly contribute to the realness of generated contents, these three aspects are widely investigated by the community. Traditional real-world simulation methods have long relied on graphics techniques that incorporate geometry, texture, and dynamics. Specifically, geometry and texture modeling~\cite{catmull19803} are employed to create the objects, while methods like keyframe animation~\cite{burtnyk1971computer} and physics-based simulation~\cite{erleben2005physics} are adopted to simulate the movement and behavior of objects over time. Despite great progress, these traditional methods often require extensive manual designs, heuristic rule definitions, and computationally expensive processing, limiting their scalability and adaptability to diverse scenarios. Recently, learning-based approaches, particularly multimodal generative models, have revolutionized content creation by providing a data-driven approach to realistic simulations. These approaches reduce the reliance on manual efforts, improve generalization across tasks, and enable intuitive interactions between humans and models. For example, Sora~\cite{videoworldsimulators2024} has garnered significant attention for its realistic simulation capabilities, demonstrating an early-stage understanding of physical laws. The emergence of such generative models introduces new perspectives and methodologies, addressing the limitations of traditional methods by reducing the need for extensive manual design and computationally expensive modeling while enhancing adaptability and scalability in diverse scenarios.

\begin{figure}[!t]
  \centering
  \includegraphics[width=0.95\linewidth]{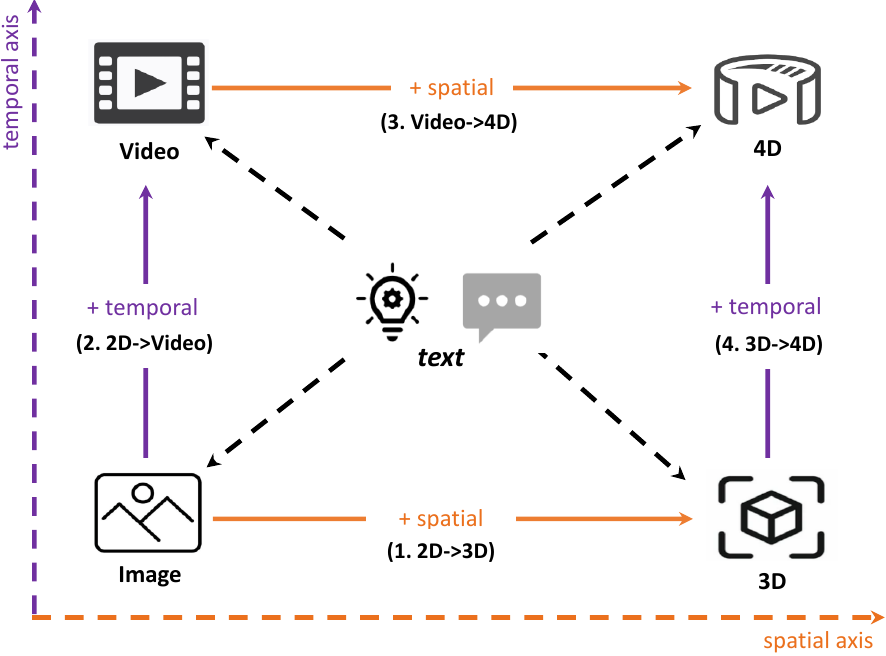}
  \caption{\textbf{The Dimensional Evolution of Generative AI.} We present a unified framework connecting 2D, Video, 3D, and 4D generation through text-guided synthesis. This paradigm illustrates how higher-dimensional content is synthesized by extending foundational modalities along spatial and temporal axes.
(1) \textit{2D$\to$3D}~\cite{wang2023score,poole2024dreamfusion,lin2023magic3d,wang2024prolificdreamer}: Spatial lifting of 2D priors to achieve geometric consistency;
(2) \textit{2D$\to$Video}~\cite{bar2024lumiere,blattmann2023stable,singer2022make}: Temporal inflation of static features to capture motion dynamics;
(3) \textit{Video$\to$4D}~\cite{jiang2023consistent4d,wu2025cat4d,wu2025sc4d,zhang20244diffusion}: Spatial reconstruction and stabilization of dynamic sequences;
(4) \textit{3D$\to$4D}~\cite{singer2023text,bah20244dfy,ren2023dreamgaussian4d,yu20244real}: Temporal animation and deformation of static geometry.
This perspective underscores that higher-dimensional generation methodologies are derivatives of foundational lower-dimensional generative priors, adapted through specialized architectural extensions.}
  \label{fig:overview}
\end{figure}

Though existing generative models offer powerful techniques for synthesizing realistic content in distinct data dimensions, the real world exhibits inherently high-dimensional complexity, and a comprehensive review that systematically integrates these advancements across different dimensions is still absent. This survey aims to bridge this gap by unifying the study of real-world simulation from the perspective of data dimension growth, as illustrated in Fig.~\ref{fig:pipeline}. Specifically, we start from 2D generation (appearance only) and then extend it to video and 3D generation by incorporating dynamics and geometry dimensions, respectively. 
Finally, we culminate in 4D generation by integrating all dimensions. To further clarify the conceptual and methodological relationships among these modalities, we provide a unified schematic in Fig.~\ref{fig:overview}.

In summary, this survey makes three key contributions. \textit{First}, it provides a systematic review of methods for real-world simulation from the perspective of data dimension growth through the lens of multimodal generative models. To the best of our knowledge, this is the first survey that unifies the study of 2D, video, 3D, and 4D generation, offering a structured and comprehensive overview of this research area. \textit{Second}, it surveys the commonly used datasets, their properties, and the corresponding evaluation metrics from various perspectives. \textit{Third}, it identifies open research challenges, aiming to guide further exploration in this field.

Previous surveys on generation models mainly concentrate on text-to-image, text-to-video, and text-to-3D generation separately without comprehensively studying their relationships. In contrast, this survey seeks to provide a more integrated perspective on multimodal generative models by tracing how generative models have evolved from handling appearance alone (2D generation), to incorporating dynamics (video generation) and geometry (3D generation), and ultimately to integrating appearance, dynamics, and geometry in 4D generation. This dimensional perspective aims to bridge previously isolated research areas and highlight common challenges and opportunities across them.

We envision this survey as a resource that provides insights for newcomers and promotes critical analysis among experienced researchers. The remainder of the survey is structured as follows. Sec.~\ref{sec:paradigms} presents four key paradigms: 2D, video, 3D, and 4D generation. The corresponding datasets and evaluation metrics for these paradigms are detailed in Appendix~\ref{sec:eval}. Finally, Sec.~\ref{sec:direction} discusses potential future research directions, and Sec.~\ref{sec:conclusions} concludes the survey. 
For non-specialists in this field, comprehensive definitions of technical terms and concepts are provided in Appendix~\ref{sec:glossary}.
Due to page limitations, the preliminaries are provided in Appendix~\ref{sec:background}.

\section{Paradigms}
\label{sec:paradigms}

This section presents methods for simulating the real world from the perspective of data dimension growth. It begins with 2D generation (Sec.~\ref{sec:2D}) for appearance modeling and then move to video generation (Sec.~\ref{sec:video}) and 3D generation (Sec.~\ref{sec:3d}) by incorporating dynamics and geometry dimensions. Finally, by integrating all these three dimensions, recent advances in 4D generation (Sec.~\ref{sec:4d}) are presented. 

\subsection{2D Generation}
\label{sec:2D}

Recently, significant advancements have been made in the field of generative models, particularly in text-to-image generation~\cite{bie2024renaissance,zhan2023multimodal,yin2025training,yin2025consistedit}. Text-to-image generation has attracted attention for its capability to produce realistic images from textual descriptions by capturing the appearance of the real world. Utilizing techniques like diffusion models, large language models (LLMs), and autoencoders, these models achieve high-quality and semantically accurate image generation.

\subsubsection{Algorithms}
\label{sec:2D-algorithms}

\myPara{Imagen}~\cite{saharia2022photorealistic} builds on the principles established by GLIDE but introduces significant optimizations and improvements.
Instead of training a task-specific text encoder from scratch, Imagen uses pre-trained and frozen language models and reduces computational demands. 
Imagen tested models trained on image-text datasets (e.g., CLIP~\cite{radford2021learning}) and models trained on pure text datasets (e.g., BERT~\cite{devlin2018bert} and T5~\cite{raffel2020exploring}). 
This practice shows that scaling up language models enhances image fidelity and text congruence more effectively than enlarging image diffusion models.

\myPara{DALL-E}~\cite{ramesh2021zero} (version 1) uses a transformer architecture that processes both text and images as a single stream of data.
DALL-E 2~\cite{ramesh2022hierarchical} utilizes the powerful semantic and stylistic capabilities of CLIP~\cite{radford2021learning}, which employs a generative diffusion decoder to reverse the process of the CLIP image encoder.
DALL-E 3~\cite{dalle3} builds upon the advancements of DALL-E 2~\cite{ramesh2022hierarchical}, offering significant improvements in image fidelity and text alignment. It enhances text understanding, allowing for more accurate and nuanced image generation from complex descriptions. DALL-E 3 is integrated with ChatGPT~\cite{chatgpt}, enabling users to brainstorm and refine prompts directly within the ChatGPT interface, which simplifies the process of generating detailed and tailored prompts. The model produces images with higher realism and better alignment to the provided text, making it a powerful tool for both creative and professional applications.

\myPara{DeepFloyd IF}~\cite{deepfloydif} is celebrated for its exceptional photorealism and advanced language understanding. This system is modular, featuring a static text encoder and three sequential pixel diffusion modules. Initially, the base model creates 64$\times$64 pixel images from textual descriptions. These images are then enhanced to 256$\times$256 pixels and further to 1024$\times$1024 pixels by two super-resolution models. Each phase utilizes a static text encoder derived from the T5~\cite{raffel2020exploring} transformer to generate text embeddings, which are subsequently processed by a U-Net architecture with integrated cross-attention and attention pooling mechanisms.

\myPara{Stable Diffusion} (SD)~\cite{rombach2022high}, also known as Latent Diffusion Model (LDM), enhances training and inference efficiency on limited computational resources while producing high-quality and diverse images.
The denoising process takes place in the latent space of pre-trained autoencoders, which map images into a spatial latent space. The underlying U-Net architecture is augmented with a cross-attention mechanism to model the conditional distribution, which can include text prompts, segmentation masks, and more. 
It used CLIP~\cite{radford2021learning} text embeddings as condition and trained on the LAION~\cite{schuhmann2021laion} dataset to generate images at a resolution of 512$\times$512 (with a latent resolution of 64$\times$64).
Building on Stable Diffusion, SDXL~\cite{podell2023sdxl} employs a U-Net backbone that is three times larger. It introduces additional attention blocks and a larger cross-attention context by utilizing a second text encoder. Additionally, SDXL includes a refinement model that enhances the visual fidelity of samples generated by SDXL through a post-hoc image-to-image technique.

\myPara{FLUX.1}~\cite{flux} utilizes a hybrid architecture that integrates multimodal and parallel diffusion transformer blocks, achieving a remarkable scale of 12 billion parameters. By employing flow matching, a straightforward yet effective technique for training generative models, FLUX.1 outperforms prior state-of-the-art diffusion models. The suite also features rotary positional embeddings and parallel attention layers, greatly improving model performance and efficiency.

\vspace{-0.3cm}

\subsection{Video Generation} 
\label{sec:video}

\begin{figure}
    \centering
    \includegraphics[width=1\linewidth]{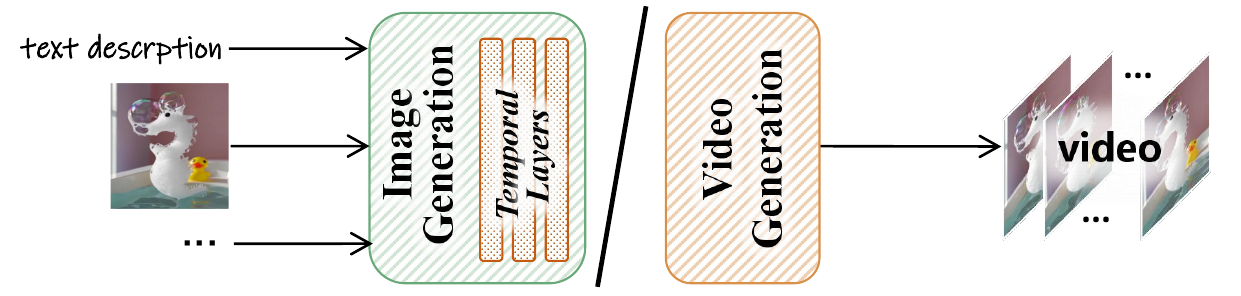}
    \caption{An illustration of the video generation paradigm. Video generation models are constructed on top of image generation models by adding temporal layers or from scratch.}
    \label{fig:relation_video_generation}
\end{figure}

Due to the structural similarities between images and videos, early approaches to video generation predominantly adapted and fine-tuned existing 2D image generation models (Sec.~\ref{sec:2D}). Early approaches addressed this challenge by adding temporal layers, such as attention or convolution, to model dynamics (Fig.~\ref{fig:relation_video_generation}), often training with mixed 2D images and video data to improve visual quality.
Inspired by Sora~\cite{videoworldsimulators2024}, state-of-the-art models now frequently employ a diffusion transformer architecture. These models operate on a compressed spatio-temporal latent space by breaking down video into a series of ``patches'' that serve as tokens for the transformer. This approach processes both spatial and temporal information simultaneously. Despite their sophistication, these models typically retain the ability to process single images as one-frame videos, allowing them to leverage the vast data available for 2D image generation. In this section, we classify these models into three main categories based on their underlying generative machine learning architectures. Fig.~\ref{fig:text-to-video} summarizes recent text-to-video methods. Readers are referred to another survey~\cite{singh2023survey} for an in-depth exploration.

\subsubsection{Algorithms}
\label{sec:t2vand2d2v}

\textbf{(1) VAE- and GAN-based Approaches.}
Before diffusion models, video generation research has advanced through two primary approaches: VAE-based and GAN-based methods, each contributing unique solutions to the challenges of video synthesis.
VAE-based methods evolved from SV2P~\cite{babaeizadeh2017stochastic} in stochastic dynamics to a combination of VQ-VAE~\cite{van2017neural} with transformers in VideoGPT~\cite{yan2021videogpt}, efficiently handling high-resolution videos through hierarchical discrete latent variables.
Notable improvements came from the parameter-efficient architecture in FitVid~\cite{babaeizadeh2021fitvid} and the integration of adversarial training for more realistic predictions.
Parallel developments in GAN-based approaches brought significant innovations, starting with MoCoGAN~\cite{mocogan}, which decomposes content and motion components for controlled generation. StyleGAN-V~\cite{skorokhodov2022stylegan} advances this by treating videos as time-continuous signals through positional embeddings, while DIGAN~\cite{yu2022generating} introduces implicit neural representations for improved continuous video modeling. StyleInV~\cite{wang2023styleinv} leverages a temporal style-modulated inversion network with pre-trained StyleGAN~\cite{karras2019style} generators, marking another milestone in high-quality frame synthesis with temporal coherence.

\begin{figure}[t]
    \centering
    \includegraphics[width=0.9\linewidth]{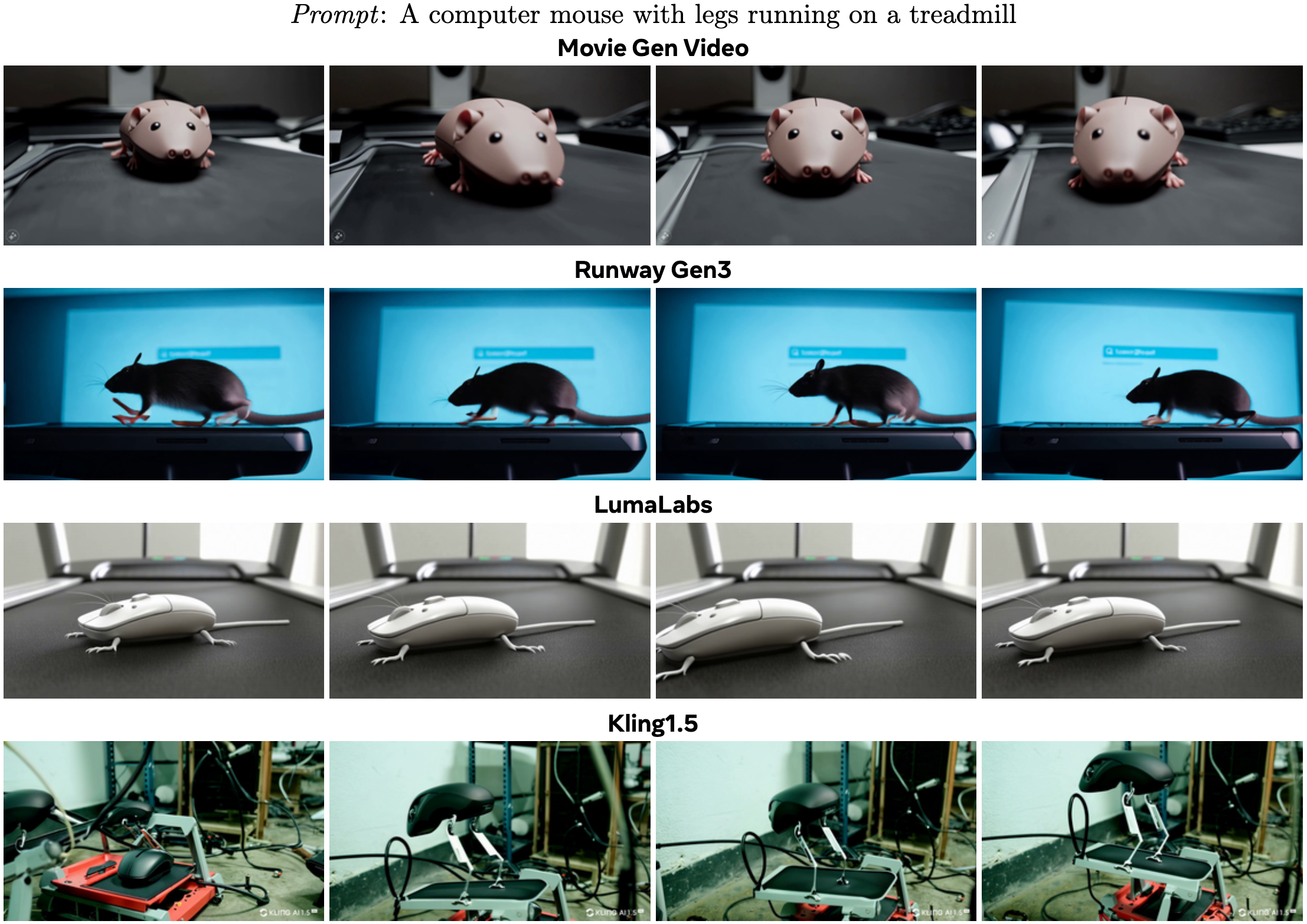}
    \caption{Qualitative comparison between different video generation methods. Results are obtained from {Movie Gen}~\cite{polyak2024movie}.}
    \label{fig:video_qualitative_comparison}
\end{figure}

\myPara{(2) Diffusion-based Approaches.} Text-to-video generation has advanced rapidly, with approaches generally falling into two categories: U-Net-based architectures and transformer-based architectures.
Fig.~\ref{fig:video_qualitative_comparison} and Table~\ref{tab:comparison_text_to_video} present qualitative results and quantitative comparisons, respectively.

\textbf{(i)	U-Net-based Architectures.}
The pioneering Video Diffusion Models (VDM)\cite{ho2022video,wang2025universe} achieved high-fidelity, temporally coherent video generation by extending image diffusion architectures and introducing joint image-video training for reduced gradient variance. Make-A-Video~\cite{singer2022make} advanced text-to-video generation without paired text-video data by leveraging existing visual representations\cite{radford2021learning} and innovative spatial-temporal modules. Imagen Video~\cite{ho2022imagen} introduced a cascade of diffusion models combining base generation with super-resolution, while MagicVideo~\cite{zhou2022magicvideo} achieved efficient generation through latent diffusion in low-dimensional space.
GEN-1~\cite{esser2023structure} focused on structure-preserving editing using depth estimates, while PYoCo~\cite{ge2023preserve} demonstrated efficient fine-tuning with limited data through carefully designed video noise priors. Align-your-Latents~\cite{blattmann2023align} achieved high-resolution generation ($1280 \times 2048$) by extending Stable Diffusion~\cite{rombach2022high} with temporal alignment techniques. Show-1~\cite{zhang2024show} combined pixel-based and latent-based approaches for enhanced quality and reduced computation.
VideoComposer~\cite{wang2024videocomposer} introduced a novel paradigm for controllable synthesis through a Spatio-Temporal Condition encoder, enabling flexible composition based on multiple conditions. AnimateDiff~\cite{guo2023animatediff} presented a plug-and-play motion module with transferable motion priors and introduced MotionLoRA for efficient adaptation. PixelDance~\cite{zeng2024make} enhanced generation by incorporating both first and last frame image instructions alongside text prompts. 

\textbf{(ii) Transformer-based Architectures.}
Following the success of Diffusion Transformer (DiT)\cite{peebles2023scalable}, transformer-based models gained prominence. VDT\cite{lu2023vdt} introduced modularized temporal and spatial attention mechanisms for diverse tasks, including prediction and interpolation. W.A.L.T~\cite{gupta2024photorealistic} achieved photorealistic generation through a unified latent space and causal encoder architecture, producing high-resolution videos at $512\times896$.
Snap Video~\cite{menapace2024snap} improved training efficiency 3.31 times through spatially and temporally redundant pixel handling, while GenTron~\cite{chen2024gentron} scaled to over 3B parameters with motion-free guidance. Luminia-T2X~\cite{gao2024lumina} integrated multiple modalities through zero-initialized attention and tokenized latent spatial-temporal space. CogVideoX~\cite{yang2024cogvideox} excelled in long-duration video generation through expert transformers, 3D VAEs, and progressive training, achieving state-of-the-art performance validated by multiple metrics.
The groundbreaking Sora~\cite{videoworldsimulators2024} is an advanced diffusion transformer model that emphasizes generating high-quality images and videos across different resolutions, aspect ratios, and durations. Sora achieves flexible and scalable generation capabilities by tokenizing the latent spatial-temporal space.

\myPara{(3) Autoregressive-based Approaches.} Parallel to diffusion-based methods, autoregressive frameworks inspired by large language models (LLMs) have emerged as an alternative approach to video generation. These methods typically follow a two-stage process: first encoding visual content into discrete latent tokens using vector quantized auto-encoders like VQ-GAN~\cite{esser2021taming} and MAGVIT~\cite{yu2023magvit, yu2023language, teng2024accelerating}, then modeling the token distribution in the latent space.
CogVideo~\cite{hong2022cogvideolargescalepretrainingtexttovideo}, a 9-billion-parameter transformer model built upon the pre-trained text-to-image model CogView~\cite{ding2021cogviewmasteringtexttoimagegeneration}, represents a significant advancement in this direction. It employs a multi-frame-rate hierarchical training strategy to enhance text-video alignment and, as one of the first open-source large-scale pre-trained text-to-video models, establishes new benchmarks in both machine and human evaluations.
VideoPoet~\cite{kondratyuk2023videopoet} introduces a decoder-only transformer architecture for zero-shot video generation, capable of processing multiple input modalities, including images, videos, text, and audio. Following LLM training paradigms with pretraining and task-specific adaptation stages, VideoPoet achieves state-of-the-art performance in zero-shot video creation, particularly excelling in motion fidelity through its diverse generative pretraining objectives.

\textbf{Evaluation.}
The evaluation of video generation models has evolved with the task's increasing complexity. Early approaches relied on distribution-based metrics, most notably the Fréchet Video Distance (FVD)~\cite{unterthiner2018towards, blattmann2023align, ho2022video}. As a temporal extension of the Fréchet Inception Distance (FID)~\cite{heusel2017gans}, FVD compares spatiotemporal feature distributions to assess visual quality and coherence. More recent benchmarks, such as VBench~\cite{huang2024vbench}, offer more granular analysis of specific attributes like motion smoothness and subject identity using features from models like CLIP~\cite{radford2021learning} and DINO~\cite{caron2021emerging}. 
However, these automated metrics often misalign with human perception.
For instance, FVD is useful for catching spatiotemporal artifacts but is sensitive to video length and resolution, largely ignoring instruction following and identity consistency.
Scores based on CLIP, DINO capture text–video alignment and subject consistency, yet provide little about temporal stability or physical plausibility, so multi-template prompts and simple robustness checks are recommended.
Suites like VBench offer useful diagnostics but can encourage overfitting, thereby highlighting the importance of cross-benchmark validation and rank consistency.

To mitigate the limitations of the automatic metrics, the field has increasingly shifted towards human studies for a more holistic and accurate assessment, particularly for advanced open-domain models.
Table~\ref{tab:comparison_text_to_video} presents a human preference evaluation of modern video generation models. 
Looking ahead, there is a high demand for reliable measures for long-horizon coherence and narrative structure, for compositional alignment across subject, action, and scene, as well as for scalable tests of physical and causal plausibility.

\begin{table}[t]
    \centering
    \scalebox{0.9}{
    \begin{tabular}{l|c|c|c|c}
        \hline
        Dimension~($\uparrow$) & Hunyuan~\cite{kong2024hunyuanvideo} & Mochi~\cite{genmo2024mochi} & Sora~\cite{videoworldsimulators2024} & Wan-14B~\cite{wan2025wan} \\
        \hline
        Motion Quality & 0.413 & 0.420 & 0.482 & 0.415 \\
        Human Artifacts & 0.734 & 0.622 & 0.786 & 0.691 \\
        Pixel Stability & 0.983 & 0.981 & 0.952 & 0.972 \\
        ID Consistency & 0.935 & 0.930 & 0.925 & 0.946 \\
        Physical & 0.898 & 0.728 & 0.933& 0.939 \\
        Smoothness & 0.890 & 0.530 & 0.930& 0.910 \\
        Image Quality & 0.605 & 0.530 & 0.665 & 0.640 \\
        Scene Quality & 0.373 & 0.368 & 0.388 & 0.386 \\
        Stylization & 0.386 & 0.403 & 0.606 & 0.328 \\
        Object Accuracy & 0.912 & 0.949 & 0.932& 0.952 \\
        \hline
    \end{tabular}}
    \caption{Quantitative comparison between different text-to-video methods. Results are obtained from~\cite{wan2025wan}.}
    \label{tab:comparison_text_to_video}
\end{table}

\subsubsection{Applications}
\label{sec:Video-apps}
\textbf{(1) Video editing} has recently benefited significantly from diffusion models, enabling sophisticated modifications while maintaining temporal consistency. The field has evolved through several innovative approaches addressing different aspects of video manipulation.
Early developments include Tune-A-Video~\cite{wu2023tune}, which pioneered a one-shot tuning paradigm extending text-to-image diffusion models to video generation through spatiotemporal attention mechanisms.
Temporal consistency has been addressed through various approaches. VidToMe~\cite{li2024vidtome} introduced token merging for aligning frames, while EI~\cite{zhang2024towards} developed specialized attention modules.
Several works focused on specialized editing capabilities. Ground-A-Video~\cite{jeong2023ground} tackled multi-attribute editing through a grounding-guided framework, while Video-P2P~\cite{liu2024video} introduced cross-attention control for character generation. 
Recent frameworks like UniEdit~\cite{bai2024uniedit} and AnyV2V~\cite{ku2024anyv2v} represent the latest evolution, offering tuning-free approaches and simplified editing processes. Specialized applications such as CoDeF~\cite{ouyang2024codef} and Pix2Video~\cite{ceylan2023pix2video} have introduced innovative techniques for temporally consistent processing and progressive change propagation.
These methods successfully balance content editing with structural preservation, marking significant advancements in video manipulation technology. \textbf{(2) Novel view synthesis} has been revolutionized by video diffusion models, which benefit from learned priors on real-world geometry, enabling high-quality view generation from limited input images.
ViewCrafter~\cite{yu2024viewcrafter} pioneered this direction by integrating video diffusion models with point-based 3D representations, introducing iterative synthesis strategies and camera trajectory planning for high-fidelity results from sparse inputs.
Camera control has emerged as a crucial aspect, with CameraCtrl~\cite{he2024cameractrl} introducing precise camera pose control through a plug-and-play module. 
Several methods have studied view consistency problems. ViVid-1-to-3~\cite{kwak2024vivid} reformulated novel view synthesis as video generation of camera movement, while NVS-Solver~\cite{you2024nvs} introduced a zero-shot paradigm that modulates diffusion sampling with given views. 
This trend shows a convergence toward leveraging video diffusion priors while maintaining geometric consistency and camera control, enabling increasingly realistic synthesis applications. \textbf{(3) Human animation in videos} has gained significance in video generation, which plays a pivotal role in the world simulator. This is particularly important because humans are the most essential participants in the real world, making their realistic simulation crucial.
Building on early generative model successes, several works~\cite{mocogan} use GANs~\cite{goodfellow2020generative} to animate humans in videos. However, maintaining high visual fidelity remains the central challenge. ControlNet~\cite{controlnet} and HumanSD~\cite{humansd} are plug-and-play methods for animating humans referring to poses based on a foundation text-to-image model, like Stable Diffusion~\cite{rombach2022high}. Additionally, to resolve the generalization issue of these methods, animate-anyone~\cite{animateanyone} proposes a ReferenceNet to maintain more spatial details of the reference video and pushes generation quality in the wild to a new milestone. Follow-up works~\cite{zeroshotha} simplify Animate-Anyone’s training and costs. Meanwhile, studies in graphics introduce 3D modeling to human video animation, with Liquid Warping GAN~\cite{liquidgan}, CustomHuman~\cite{custom_humans}, and LatentMan~\cite{LatentMan} being early efforts to integrate 3D human priors into the generation loop. The latest progress, MIMO~\cite{mimo}, explicitly models the charter, 3D motion, and scene respectively to drive human animation in the wild. These methods make a great step toward introducing humans into the loop of the world simulator.

\begin{figure*}[!t]
	\centering
	\includegraphics[width=1\textwidth]{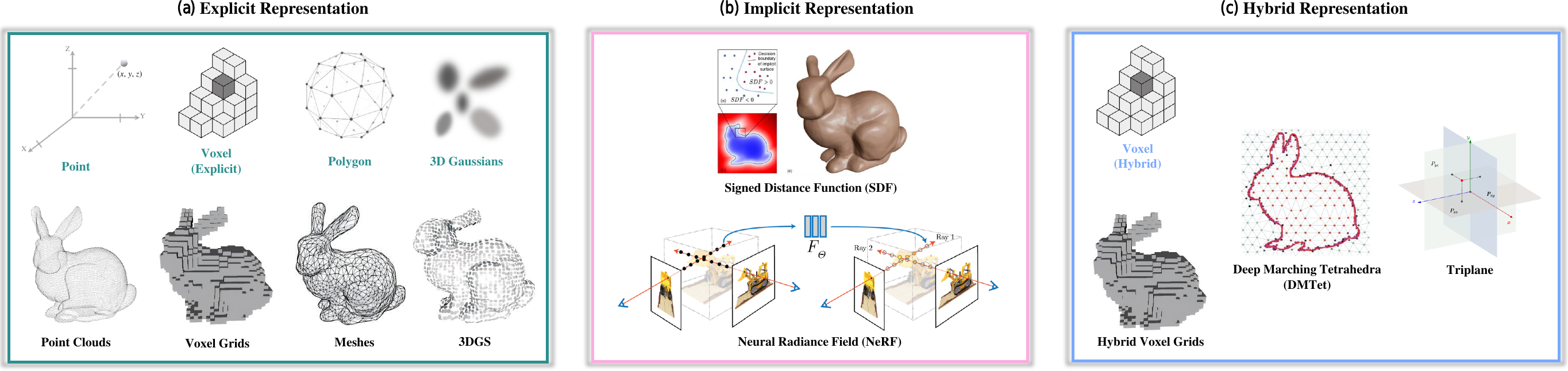}
	\caption{Three main categories of neural scene representations. (a) Explicit representation stores geometry directly using point clouds, voxel grids~\cite{gupta20203d}, meshes~\cite{rossi2021robust}, and 3D Gaussians~\cite{kerbl20233d}. (b) Implicit representation defines objects through functions like Signed Distance Functions (SDF)~\cite{park2019deepsdf} and Neural Radiance Fields (NeRF)~\cite{mildenhall2020nerf}, enabling smooth, continuous surfaces without fixed resolution. (c) Hybrid representation combines explicit and implicit methods, using techniques like Hybrid Voxel Grids, Deep Marching Tetrahedra (DMTet)~\cite{shen2021dmtet}, and Triplanes for better efficiency and flexibility.}
	\label{3d_representation}
\end{figure*}
\subsection{3D Generation}
\label{sec:3d}

\begin{figure*}[t]
	\centering
	\includegraphics[width=1\textwidth]{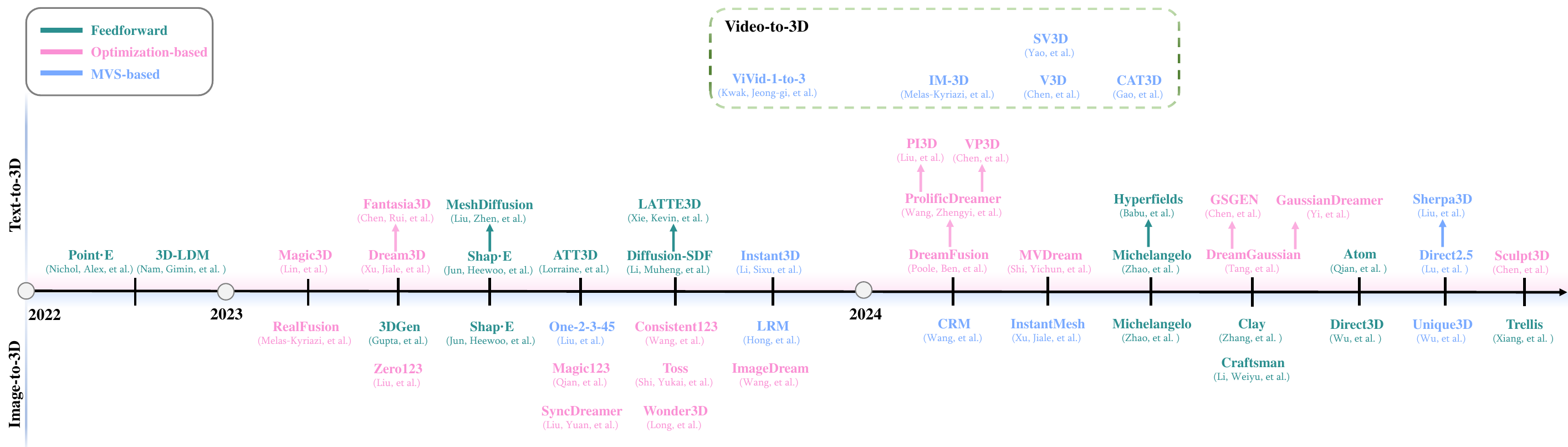}
	\caption{Chronological overview of recent text-to-3D, image-to-3D and video-to-3D generation methods.}
	\label{3d_chrono}
\end{figure*}

3D generation focuses on both geometry and appearance to better simulate real-world scenarios. In this section, we explore various 3D representations and generation algorithms, providing a structured overview of recent advancements. Specifically, we categorize 3D generation methods based on their input modalities, including \textit{Text-to-3D Generation}, which directly synthesizes 3D content from textual descriptions, \textit{Image-to-3D Generation}, which introduces image constraints to refine text-driven outputs, and \textit{Video-to-3D Generation}, which leverages video priors for more consistent 3D generation. A chronological summary of these advancements is presented in Fig.~\ref{3d_chrono}, while Table~\ref{tab:text-to-3d} offers a comprehensive comparison of cutting-edge methods. Notably, several approaches span multiple categories, demonstrating the versatility of modern 3D generation techniques.

Instead of constructing 3D generation models from scratch, most existing approaches are highly coupled with 2D and video generation models to leverage their powerful appearance modeling capability, as illustrated in Figs.~\ref{fig:relation_t23d}, \ref{fig:relation_i23d}, and \ref{fig:relation_v23d}. \textit{First}, the image prior encoded in 2D and geometry cues encoded in video generation models can be employed to provide supervision for 3D generation models. \textit{Second}, 2D and video generation models can be fine-tuned to take additional 3D information (e.g., normals) as input to synthesize 3D-aware multi-view images to facilitate 3D generation. 

\subsubsection{3D Representation}
\label{sec:3d-rep}

In the domain of 3D generation, the choice of an optimal 3D representation is crucial. For neural scene representation, 3D data can be typically divided into three main categories: explicit, implicit, and hybrid representations, which are illustrated in Fig.~\ref{3d_representation}.

\noindent{\textbf{(1) Explicit Representations.}}
Explicit representations offer precise visualizations of objects and scenes, defined by a set of elements. Traditional forms, such as point clouds, meshes, and voxels, have been widely used for years.

\textbf{(i)	Point Clouds} are unordered sets of 3D points $(x, y, z)$ and often enriched with attributes like color or normals. While widely used due to depth sensor output, their irregular structure~\cite{liu2019deep, qi2017pointnet} challenges conventional neural networks in 2D domains. \textbf{(ii) Voxel Grids} are composed of basic elements called voxels. A voxel, as a 3D counterpart of a pixel \cite{blinn2005pixel}, represents points on a 3D grid. It subdivides a bounding box into smaller elements, each with an occupancy value. While voxels can store data like opacity or color, they are memory-inefficient for high-resolution data due to cubic memory growth ($\mathcal{O}(n^3)$) and sparse occupancy. \textbf{(iii) Meshes} composed of vertices and edges forming polygons (faces), define two-dimensional surfaces in three-dimensional space. They are more memory-efficient than voxel grids by encoding only object surfaces and, unlike point clouds, provide explicit connectivity, enabling geometric transformations and efficient texture encoding. \textbf{(iv) 3D Gaussian Splatting (3DGS)} is proposed as an effective method for accelerating both training and rendering tasks~\cite{kerbl20233d}. This technique models objects as collections of anisotropic Gaussian distributions, where each distribution is defined by its position (i.e., mean $\mathbf{x} \in \mathbb{R}^3$), covariance matrix $\mathbf{\Sigma}$, opacity $\alpha \in \mathbb{R}$, and spherical harmonics coefficients $\mathcal{C} \in \mathbb{R}^k$ (with $k$ representing the degrees of freedom), allowing for the modeling of view-dependent color variations, which is expressed as: 
\vspace{-0.05cm}
\begin{equation}
G(X)=e^{-\frac{1}{2} \mathbf{x}^T \mathbf{\Sigma}^{-1} \mathbf{x}}.
\label{eq:gs1}
\end{equation}
To facilitate optimization, covariance matrix $\bf \Sigma$ is factorized into a scaling matrix $\mathbf{S}$ and a rotation matrix $\mathbf{R}$, such that: 

\begin{equation}
    \Sigma=\mathbf{R S S}^T \mathbf{R}^T.
\end{equation}

\noindent{\textbf{(2) Implicit Representations.}} 
Implicit representations describe 3D spaces with continuous functions, such as mathematical models or neural networks, capturing volumetric properties rather than surface geometry. Implicit neural representations approximate these functions with neural networks, enhancing expressiveness at the cost of higher training and inference overhead. Key approaches include Signed Distance Field (SDF) \cite{park2019deepsdf} and Neural Radiance Field (NeRF) \cite{mildenhall2020nerf}.

\textbf{(i)	Signed Distance Field (SDF)} is used to represent 3D shapes via neural fields, where a surface is implicitly defined as the zero-level set of the SDF. Given a point \( \mathbf{x} \in \mathbb{R}^3 \), the SDF function \( f(\mathbf{x}) = d \) returns the shortest distance to the nearest surface point, with the sign indicating whether \( \mathbf{x} \) lies inside (\( d < 0 \)) or outside (\( d > 0 \)) the shape. The surface is defined by the set of points where \( f(\mathbf{x}) = 0 \). The accuracy of SDFs relies on precise normal vectors, as their orientation determines distance signs, making accurate normal estimation essential for surface representation. \textbf{(ii) Neural Radiance Field (NeRF)} ~\cite{mildenhall2020nerf} represents 3D scenes as continuous volumetric functions encoded in a Multi-Layer Perceptron (MLP). Unlike DeepSDF ~\cite{park2019deepsdf}, NeRF regresses density and color instead of a signed distance function, allowing for volumetric rendering. Specifically, NeRF maps spatial positions \( \mathbf{x} \in \mathbb{R}^3 \) and viewing directions \( \mathbf{d} \in \mathbb{R}^2 \) to density \( \sigma \) and color \( c \), \textit{i.e.}, \( f(\mathbf{x}, \mathbf{d}) = (\sigma, c) \), using an MLP trained on images with known camera poses. To render a pixel, NeRF casts a ray \( \mathbf{r}(t) = \mathbf{o} + t\mathbf{d} \) from the camera into the scene and samples points along the ray at intervals \( t_i \). The final pixel color \( C(\mathbf{r}) \) is computed via volume rendering as:
\begin{equation}
     C(\mathbf{r}) = \sum_{i} T_i \alpha_i \mathbf{c}_i,\quad 
     \text{where}\  T_i = \exp \left( -\sum_{k=0}^{i-1} \sigma_k \delta_k \right),
    \label{eq:vol_ren} 
\end{equation}
the term \( \alpha_i = 1 - \exp(-\sigma_i \delta_i) \) here represents the opacity of the sampled point, and the accumulated transmittance \( T_i \) quantifies the probability that the ray travels from \( t_0 \) to \( t_i \) without being occluded by any particles. The term \( \delta_i = t_i - t_{i-1} \) denotes the distance between consecutive sampled points along the ray.

\noindent{\textbf{(3) Hybrid Representations.}}
Most current implicit methods depend on regressing NeRF or SDF values, which can limit their capacity to leverage explicit supervision on target views or surfaces. Explicit representations, however, offer useful constraints during training and improve user interaction. To leverage the complementary strengths of both paradigms, hybrid representations can be seen as a trade-off between explicit and implicit representations.

\textbf{(i) Hybrid Voxel Grids} can be used as hybrid representations in methods like~\cite{fridovich2022plenoxels, sun2022direct, muller2022instant}. \cite{sun2022direct} employs density and feature grids for radiance field reconstruction, while Instant-NGP~\cite{muller2022instant} uses hash-based multi-level grids, optimizing GPU performance for faster training and rendering. \textbf{(ii) Deep Marching Tetrahedra (DMTet)} ~\cite{shen2021dmtet} combines tetrahedral grids with implicit SDF for flexible 3D surface representation. A neural network predicts SDF values and position offsets for each vertex, allowing the modeling of complex topologies. The grids are converted into meshes via a differentiable Marching Tetrahedra (MT) layer, enabling efficient, high-resolution rendering. By optimizing geometry and topology with mesh-based losses, DMTet achieves finer details, fewer artifacts, and outperforms previous methods in conditional shape synthesis from coarse voxels on complex 3D datasets. \textbf{(iii) Tri-plane} provides a memory-efficient alternative to voxel grids for 3D shape representation and neural rendering.It decomposes 3D volumes into three orthogonal 2D feature planes $(XY, XZ, YZ)$. EG3D~\cite{chan2022efficient} utilizes this structure, employing an MLP to aggregate features from the planes and predict color and density values at any 3D point: \( (\sigma, c) = \texttt{MLP}\big(\textbf{f}_{xy}(\mathbf{x}) + \textbf{f}_{xz}(\mathbf{x}) + \textbf{f}_{yz}(\mathbf{x})\big) \). This approach reduces memory consumption compared to voxel-based NeRF and enables faster rendering.

\subsubsection{Algorithms} \label{sec:3d-algorithms}

\begin{figure}[t]
    \centering
    \includegraphics[width=1\linewidth]{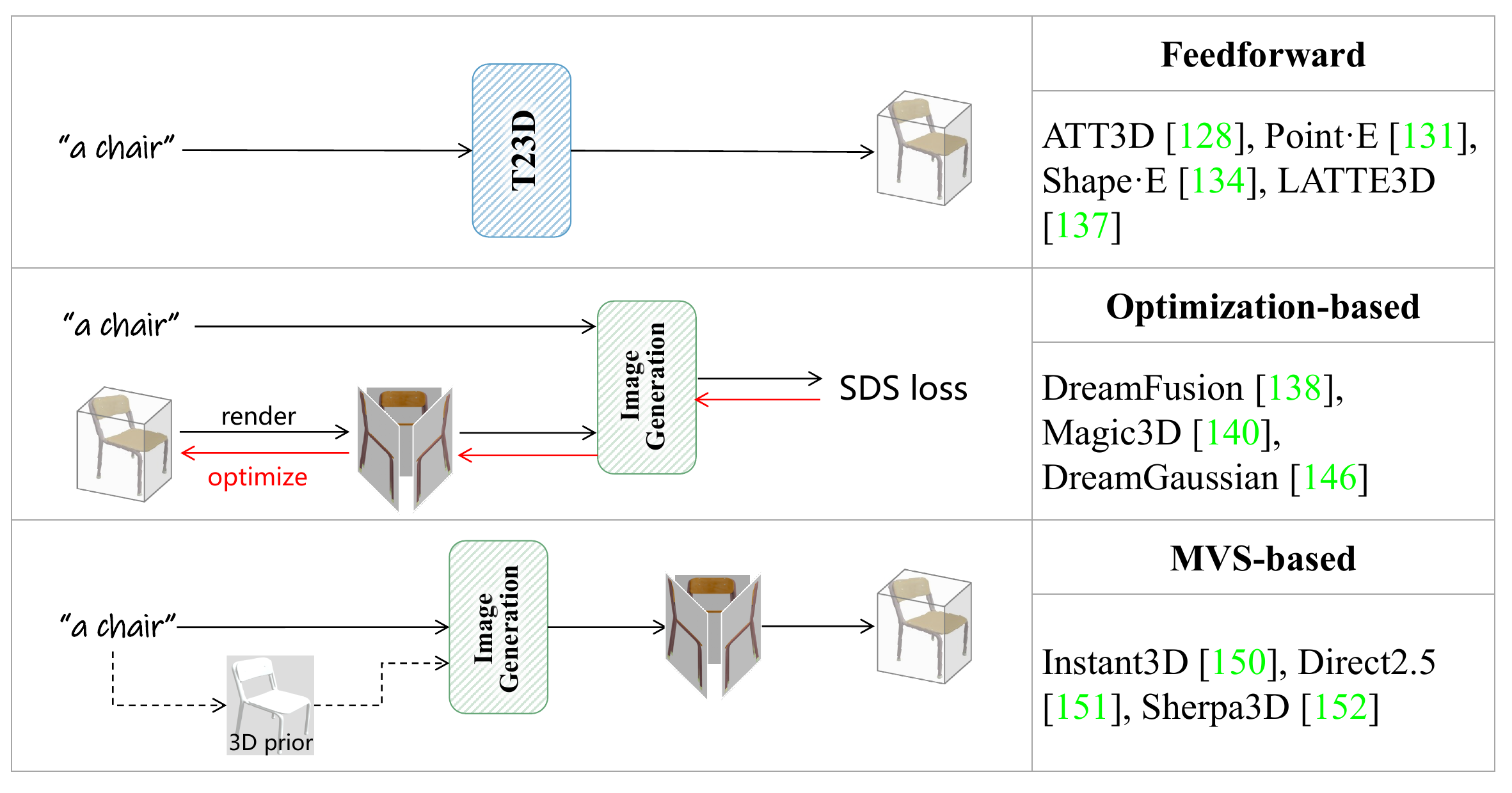}
    \caption{Comparison between different text-to-3D generation paradigms. Image generation models serve as critical components to construct text-to-3D approaches.}
    \label{fig:relation_t23d}
\end{figure}

\noindent{\textbf{(1) Text-to-3D Generation.}} 
To generate 3D content from the text prompt by simulating the geometrics in the real world, extensive studies have been conducted and can be divided into three branches. Readers can refer to \cite{li2024advances} for a more comprehensive survey in this field. Comparison of different branches of methods is illustrated in Fig.~\ref{fig:relation_t23d}. As we can see, image generation models serve as critical components in text-to-3D approaches to provide supervision (i.e., Score Distillation Sampling (SDS) loss) or synthesize multi-view images for more accurate 3D generation.

\textbf{(i) Feedforward Approaches.} Motivated by text-to-image generation, a primary branch of methods extends existing generative models to directly synthesize 3D representations from the text prompt in one feedforward propagation. The key to success lies in encoding the 3D geometry into a compact representation and aligning it with the corresponding text prompt.

Michelangelo \cite{zhao2023michelangelo} first constructs a VAE model to encode 3D shapes into a latent embedding. Then, this embedding is aligned with the ones extracted from language and image using CLIP \cite{radford2021learning}. Using a contrastive loss for optimization, a 3D shape can be inferred from the text prompt. ATT3D \cite{lorraine2023att3d} uses the Instant-NGP model as the 3D representation and bridges it with the text embedding using a mapping network. Then, multi-view images are rendered from the Instant-NGP model, and the whole network is optimized using SDS loss. Motivated by ATT3D, Atom \cite{qian2024atom} learns to predict a triplane representation from the text embedding and employs a two-stage optimization strategy. Hyperfields \cite{babu2024hyperfields} trains a dynamic hypernet to record the NeRF parameters learned from diverse scenes. Recently, the impressive performance of diffusion models has motivated researchers to extend them to 3D generation. Early methods focus on learning to synthesize explicit 3D representations from the text prompt. Specifically, Point$\cdot$E \cite{nichol2022point} first employs GLIDE \cite{nichol2022glide} to synthesize multiple views, which are then used as conditions to produce a point cloud using a diffusion model. Later, MeshDiffusion \cite{liu2023meshdiffusion} employs diffusions to build a mapping from text to meshes. Subsequent methods make attempts to apply diffusion models to implicit 3D representations. Shap$\cdot$E \cite{shape-e} first maps 3D contents to parameters of a radiance field and then trains a diffusion model to generate these parameters conditioned on the text embedding. 3D-LDM~\cite{nam20223d} represents geometry using SDFs and trains a text-conditioned diffusion model. Diffusion-SDF~\cite{li2023diffusion} extends this idea with an SDF autoencoder and voxel-based diffusion to synthesize voxelized SDFs from text. LATTE3D~\cite{xie2024latte3d} further separates geometry and texture generation, employing dedicated networks to produce SDF- and NeRF-based outputs from text embeddings. 
Trellis \cite{xiang2025structured} introduces unified structured latent representation and develops rectified flow transformers for feedforward 3D generation.

\textbf{Discussion.} Compared with optimization-based approaches, feedforward approaches favor high efficiency and are capable of generating 3D contents without test-time optimization. These approaches, however, rely heavily on the quantity of data and usually suffer from inferior structure and texture details.

\textbf{(ii) Optimization-based Approaches.} On top of text-to-image generation, another branch of methods optimize 3D representations by leveraging powerful text-to-image generative models to provide rich supervision.  

DreamFusion \cite{poole2024dreamfusion} first introduces score distillation sampling (SDS) loss to optimize NeRFs using images synthesized from text prompts. MVDream \cite{shi2023mvdream} fine-tunes a multi-view diffusion model to generate multi-view images with cross-view consistency to train a NeRF to capture 3D contents. Magic3D \cite{lin2023magic3d} employs textured meshes to represent the 3D object and adopts SDS loss for optimization of the meshes. Dream3D \cite{xu2023dream3d} first generates an image from the text prompt, which is then employed to produce the 3D shape to initialize a neural radiance field. Next, a NeRF is optimized using the CLIP guidance. Fantasia3D \cite{chen2023fantasia3d} further combines DMTet and SDS loss to generate a 3D object from the text prompt. ProlificDreamer \cite{wang2024prolificdreamer} develops variational score distillation (VSD) to model the distribution of 3D representations and produces higher-quality results with rich details. To address the multi-face Janus issue, PI3D \cite{liu2024pi3d} first fine-tunes the text-to-image diffusion model to produce pseudo-images. Then, these images are adopted to generate a 3D shape using an SDS loss. VP3D \cite{chen2024vp3d} first uses a text-to-image diffusion model to generate a high-quality image from the text prompt. Then, 3D representations are optimized via SDS loss using the resultant image and the text prompt as conditions.

With remarkable advances in 3D Gaussian, it is widely studied in the field of text-to-3D generation. DreamGaussian \cite{tang2024dreamgaussian} first employs a diffusion to obtain 3D Gaussians with SDS loss being used for optimization. Then, meshes are extracted from 3D Gaussians with texture being refined to obtain higher quality content. To promote convergence, GSGEN \cite{chen2024text} and GaussianDreamer \cite{yi2024gaussiandreamer} first employ Point$\cdot$E to generate a point cloud from the text prompt to initialize the positions of Gaussians, which are then optimized to refine their geometry and appearance using SDS loss. Sculpt3D \cite{chen2024sculpt3d} introduces a 3D prior by retrieving reference 3D objects in a database, which can be seamlessly integrated with existing pipelines.

\textbf{Discussion.} Thanks to the rich knowledge in text-to-image models, optimization-based approaches produce finer details. However, these approaches require expensive per-prompt optimization and are time-consuming.

\textbf{(iii) MVS-based Approaches.} Instead of directly generating 3D representations from text prompts, to make better use of text-to-image models, numerous attempts have been made to synthesize multi-view images for 3D generation. 

Instant3D \cite{li2024instant3d} first fine-tunes a text-to-image diffusion model to generate four-view images. Then, these images are passed to a transformer to predict a triplane representation. Direct2.5 \cite{lu2024direct2} fine-tunes a multi-view normal diffusion model on 2.5D rendered and natural images. Given a text prompt, Direct2.5 first produces normal maps and optimizes them through differentiable rasterization. Then, the optimal normal maps are adopted as conditions to synthesize multi-view images. Sherpa3D \cite{liu2024sherpa3d} first employs a 3D diffusion model to generate a coarse 3D prior from the text prompt. Then, the normal map is produced and used to synthesize multi-view images with 3D coherence. 

\textbf{Discussion.} With recent advances of VLMs, lifting these 2D generative models for 3D generation by injecting 3D priors has drawn increasing interest. However, the formulation of 3D consistency and fine-tuning with a limited amount of 3D data remain open problems.

\begin{table}[t]
    \centering
    \setlength\tabcolsep{3pt} 
    \scalebox{0.9}{
    \begin{tabular}{llccc}
        \hline
         Type & Method & \textbf{Quality ($\uparrow$)} & \textbf{Alignment ($\uparrow$)} & \textbf{Time ($\downarrow$)}   \\
         \hline
         \multirow{7}{*}{\begin{tabular}[l]{@{}l@{}}Optimization-\\based\end{tabular}} & Magic3D \cite{lin2023magic3d} & 38.7 & 35.3 & \textasciitilde40 min \\ 
         & Fantasia3D \cite{chen2023fantasia3d} & 29.2 & 23.5 & \textasciitilde45 min \\
         & DreamFusion \cite{poole2024dreamfusion} & 24.9 & 24.0 & 30 min \\ 
         & ProlificDreamer \cite{wang2024prolificdreamer} & 51.1 & 47.8 & \textasciitilde240 min \\
         & DreamGaussian \cite{tang2024dreamgaussian} & 19.9 & 19.8 & \textasciitilde7 min \\
         & VP3D \cite{chen2024vp3d} & 54.8 & 52.2 & - \\
         & GaussianDreamer \cite{yi2023gaussiandreamer} & 54.0 & - & - \\
         \hline
         MVS-based& MVDream \cite{shi2023mvdream} & 53.2 & 42.3 & \textasciitilde30 min \\
         \hline
         Feedforward& Trellis \cite{xiang2025structured} & 35.6 & 21.4 & $<$1 min \\
         \hline
    \end{tabular}}
    \caption{Quantitative comparison on T3Bench (Single Object) between different text-to-3D methods.}
    \label{tab:comparison_text_to_3d}
\end{table}

\begin{figure}
    \centering
    \includegraphics[width=1\linewidth]{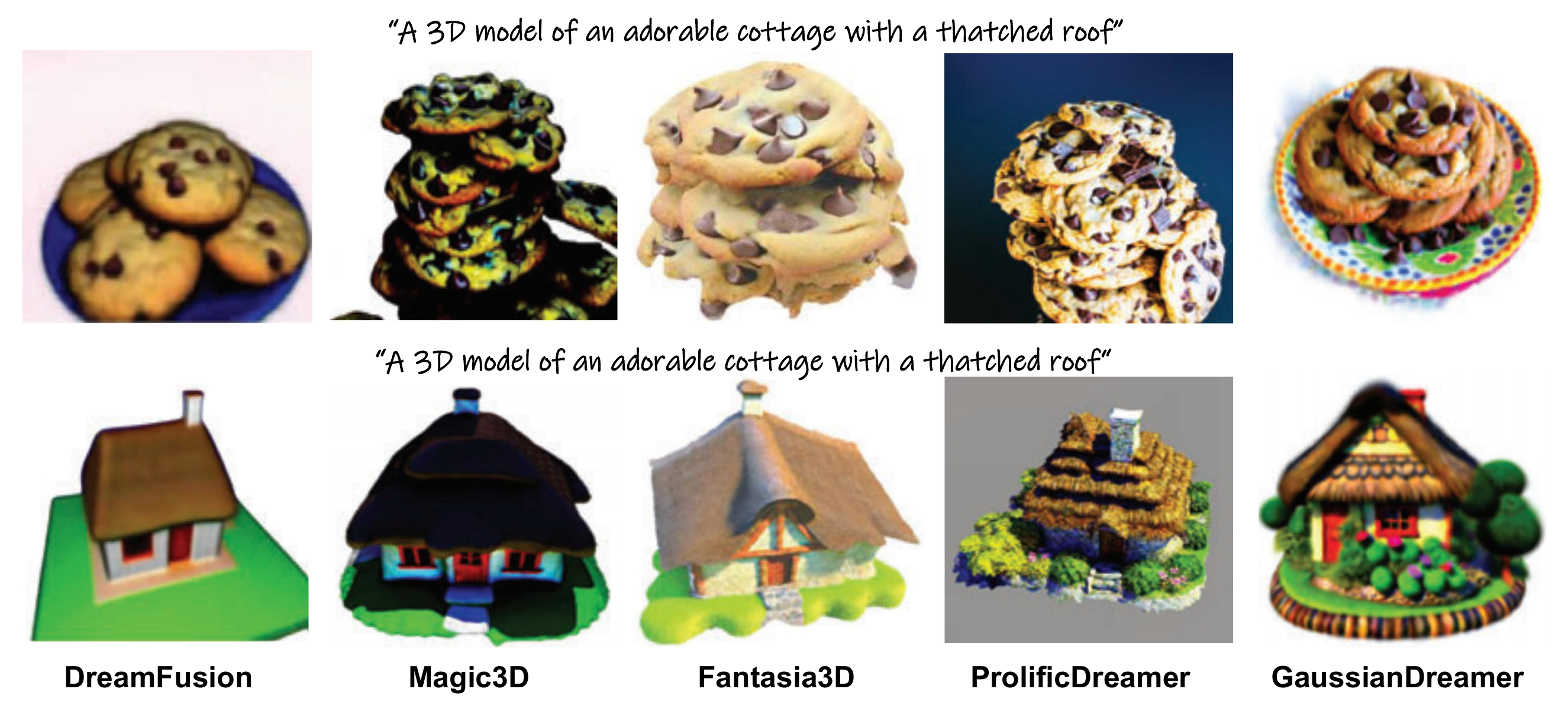}
    \caption{Comparison between different text-to-3D methods.}
    \label{fig:comparison_text_to_3d}
\end{figure}

\textbf{Evaluation.} Quantitative evaluation of text-to-3D methods remains an open problem. For subjective quality evaluation, common reference-based metrics (e.g., PSNR) are not applicable as groundtruth data are unavailable, while non-reference quality (e.g., FID) metrics may not always align with human preferences. Consequently, most methods employ CLIP Score and CLIP R-Precision to evaluate the alignment of 3D models with text prompts. Recently, several benchmarks~\cite{he2023t,su2024gt23d} have been established to comprehensively assess text-to-3D generation methods. Here, we report the quantitative scores of representative methods in Table \ref{tab:comparison_text_to_3d} and present their visual results in Fig.~\ref{fig:comparison_text_to_3d}. Readers can refer to \cite{he2023t,su2024gt23d} for more details. From these results, we summarize four key findings. (1) Quality metric focuses on geometry and subjective accuracy, while alignment metric focuses on semantic alignment with the prompt; these two metrics are not entirely correlated. For example, DreamFusion produces a higher alignment score than Fantasia3D but suffers a notable quality drop. (2) Currently, optimization-based and MVS-based approaches outperform feedforward ones by leveraging powerful image generation models at the cost of higher computational cost. (3) Feedforward approaches achieve significant speedup compared with other ones, exhibiting great potential for real-time applications. (4) The visual quality of generated 3D assets is still limited and continuous improvement is highly demanded.

\begin{figure}
    \centering
    \includegraphics[width=1\linewidth]{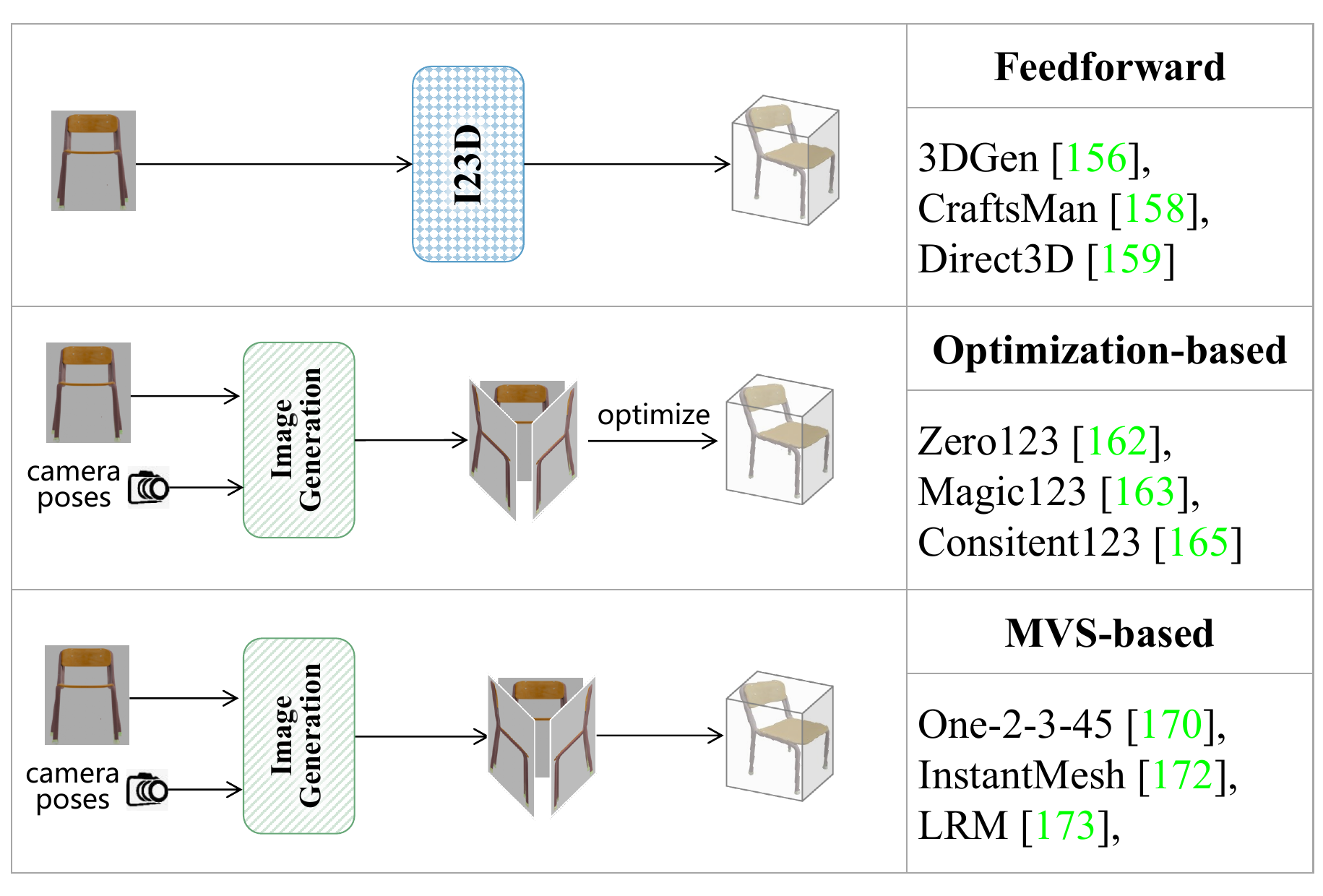}
    \caption{Comparison between different image-to-3D generation paradigms. Image generation models can be employed to synthesize multi-view images to promote 3D generation.}
    \label{fig:relation_i23d}
\end{figure}

\begin{figure}[t]
    \centering
    \includegraphics[width=0.85\linewidth]{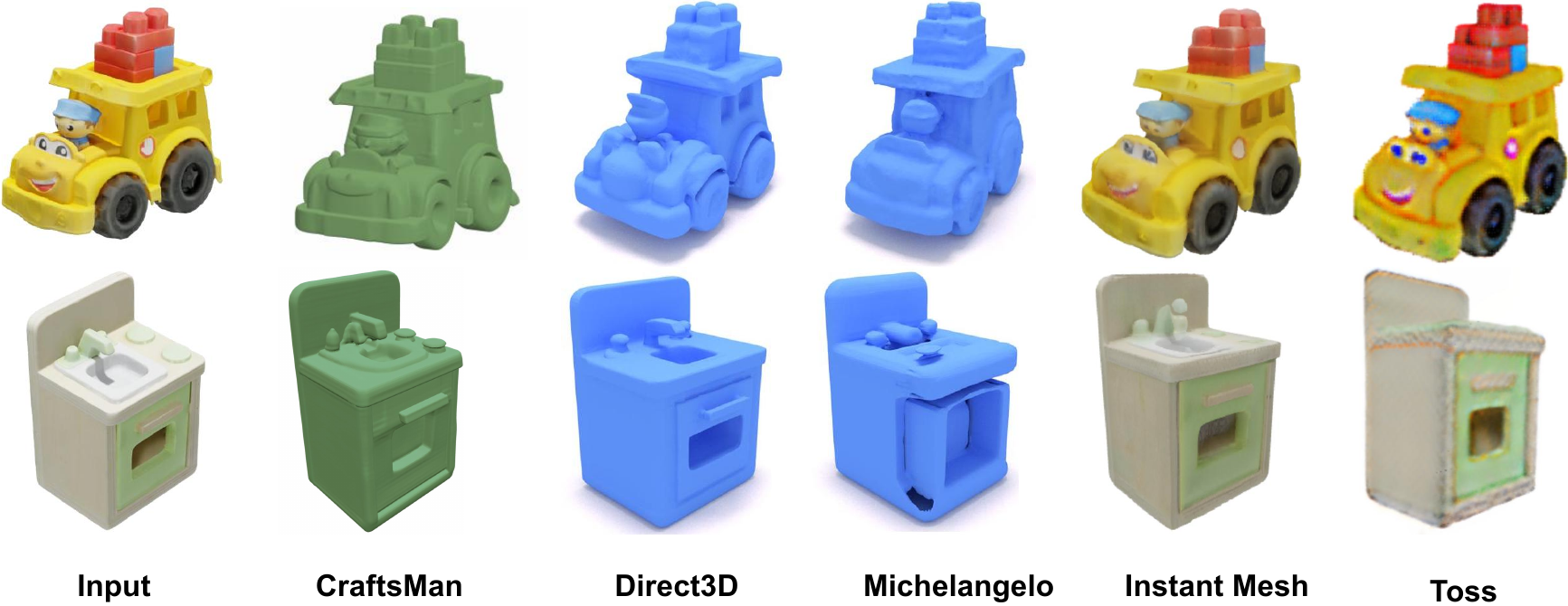}
    \caption{Comparison between different image-to-3D methods.}
    \label{fig:comparison_image_to_3d}
\end{figure}

\noindent{\textbf{(2) Image-to-3D Generation.}}
The goal of the image-to-3D task is to generate high-quality 3D assets that are consistent with the identity of the given image. Due to the high cost of 3D data collection, text-to-3D generation lacks sufficient high-quality text annotations to scale up compared with image and video generation. Since images naturally capture more low-level information that closely aligns with the 3D modality, the image-to-3D task narrows the modality gap between the input and the output compared to text-to-3D generation. Consequently, image-to-3D has emerged as a foundational task for advancing native 3D generation. To leverage the knowledge inside image generation models, they are frequently adopted as components of image-to-3D models (Fig.~\ref{fig:relation_i23d}). The qualitative comparison of the part methods is shown in Fig.~\ref{fig:comparison_image_to_3d} and the quantitative comparison is illustrated in Table~\ref{tab:image-3d quantitative results}. Due to inconsistencies in the evaluation datasets or metrics used in the paper, some works are not listed in the figure and table.

\textbf{(i) Feedforward Approaches.} These methods encode 3D assets into latent codes via a compression network (e.g., VAE) and then train a generative model on the latent space. 3DGen \cite{gupta20233dgen} uses a triplane latent space to improve compression accuracy and efficiency, while Direct3D \cite{shuang2025direct3d} applies triplane representation with direct 3D supervision to preserve fine 3D details. Michelangelo\cite{zhao2023michelangelo} draws inspiration from 3Dshape2vecset\cite{zhang20233dshape2vecset}, and utilizes 1D vectors as the latent space, when supervising the output with occupancy fields. CraftsMan\cite{li2024craftsman} further introduces a multi-view generation model to generate multi-view images as conditions of the diffusion model, followed by normal-based refinement for generated meshes. Clay\cite{zhang2024clay} brings a comprehensive system pretrained on large-scale 3D datasets for 3D generation, including VAE and diffusion models based on 1D vectors for geometry generation, Material Diffusion for PBR texture, and conditioning design across various modalities.

\textbf{Discussion.} Native methods jointly train compression and generation models on 3D datasets, achieving finer geometric details than MVS- or optimization-based approaches. However, the size of 3D datasets\cite{deitke2023objaverse,deitke2024objaverse} scales much more slowly than image or video datasets\cite{schuhmann2022laion,chen2024panda} due to the high cost of production and collection. Therefore, native methods lack sufficiently diverse and extensive data for pretraining. Consequently, how to leverage the priors from videos and images to enhance the diversity and generalization of 3D generation, especially in texture generation, remains an area for further exploration.

\textbf{(ii) Optimization-based Approaches.} With the advances of distillation-based methods in text-to-3D models, optimization-based methods optimize 3D assets through a training process supervised by SDS loss from pretrained image-to-image or text-to-image generation models, while maintaining the image identity by various additional loss constraints.

Building on DreamFusion \cite{poole2024dreamfusion}, Magic3D \cite{lin2023magic3d}, and SJC \cite{wang2023score}, RealFusion \cite{melas2023realfusion} relies on priors from pretrained text-to-image models using SDS loss, with image reconstruction and textual inversion preserving low-level and semantic identity. Leveraging large-scale 3D datasets \cite{deitke2023objaverse}, Zero123 \cite{liu2023zero} replaces the text-to-image model with a novel-view synthesis model conditioned on camera poses, combining image detail with multi-view consistency to effectively reduce the Janus problem. A line of work has expanded upon Zero123\cite{liu2023zero}. Zero123-xl\cite{deitke2024objaverse} pretrains Zero123 pipelines on a 10$\times$ larger 3D datasets for better generalization. Magic123\cite{qian2023magic123} leverages 2D and 3D priors simultaneously for distillation to manage the trade-off between generalization and consistency, and uses a coarse-fine pipeline for higher quality. SyncDreamer\cite{liu2024syncdreamer} and Consistent123\cite{weng2023consistent123} both further improve the multi-view consistency of NVS models by introducing a synchronized multi-view diffusion model, when the former leverages a 3D volume to model joint distribution relations of images and the latter utilizes cross-view attention and shared self-attention. Toss\cite{shi2023toss} additionally brings text caption as high-level semantics of 3D data into the NVS model, pretaining for stronger plausibility and controllability of invisible views. ImageDream\cite{wang2023imagedream} addresses both multi-view consistency and the 3D details problem by designing a multi-level image-prompt controller and training with text descriptions. Wonder3D\cite{long2024wonder3d} incorporates the cross-domain attention mechanism, enabling the NVS model to simultaneously denoise images and align normal maps, when introducing normal maps into the optimization process additionally. IPDreamer~\cite{zeng2023ipdreamer} enables controllable 3D synthesis from complex image prompts by introducing IPSDS, a variant of SDS, and a mask-guided alignment strategy for multi-prompt consistency.

\textbf{Discussion.} Inheriting the powerful priors of image generation models, optimization-based methods demonstrate strong generalization capabilities and can model high-precision textures. However, since novel view synthesis (NVS) models only use 2D data sampled from 3D rather than directly 3D data for supervision during pretraining, the multi-view consistency problem cannot be fundamentally resolved, despite improvements through 3D volume modeling or cross-view attention. Consequently, optimization-based methods often suffer from overly smooth geometries and long times for training due to the optimization paradigm.

\textbf{(iii) MVS-based Approaches.} MVS-based methods split the image-to-3D generation into two stages: first generate multi-view images from a single image using an NVS model, then directly create 3D assets from these multi-view images using a feedforward reconstruction network.

Based on multi-view images predicted by Zero123\cite{liu2023zero}, One-2-3-45\cite{liu2024one} proposes an elevation estimation module and utilizes an SDF-based generalizable neural surface reconstruction module pretrained on 3D datasets for $360^\circ$ mesh reconstruction, which reduces the reconstruction time to 45 seconds compared to optimization-based methods. CRM\cite{wang2024crm} further freezes the output images of the multi-view generation model to six fixed camera poses, largely improving the consistency between multi-views. Then CRM input multi-view images into a convolutional U-Net to create a high-resolution triplane supervised by depth and RGB images. InstantMesh\cite{xu2024instantmesh} also freezes the camera poses of multi-view images, but employs a transformer-based multi-view reconstruction model based on LRM\cite{hong2023lrm} to reconstruct 3D meshes, providing better generalization at the expense of some image-to-3D detail consistency. Unique3D\cite{wu2024unique3d} introduces a multi-level upscale strategy to progressively generate higher-resolution multi-view images, and a normal map diffusion model to predict multi-view normal maps for the initialization of coarse meshes, which are refined and colorized based on multi-view images.

\textbf{Discussion.} Compared with optimization-based methods, MVS-based methods train a feedforward reconstruction model on 3D datasets to reconstruct a high-quality 3D model from multi-view images, significantly improving the 3D consistency and reducing inference time to seconds level. However, MVS-based methods often lack high-quality geometry details due to the limitation of the model scale.

\textbf{Evaluation.} As shown in Table~\ref{tab:image-3d quantitative results} and Fig.~\ref{fig:comparison_image_to_3d}, existing image-to-3D methods are typically evaluated on the GSO dataset using Chamfer Distance(CD) and IoU between the generated and ground-truth geometries. We can find that the geometry quality of InstantMesh~\cite{xu2024instantmesh} and CraftsMan3D~\cite{li2024craftsman} is significantly better than that of optimization-based and other multi-view based methods. Considering that both methods are pretrained on large-scale 3D object datasets~\cite{deitke2023objaverse,deitke2024objaverse}, this highlights the crucial role of large-scale pretraining in improving geometry quality. Feed-forward and multi-view methods perform comparably, suggesting that with sufficiently large 3D datasets, multi-view inputs mainly enhance controllability rather than geometry quality. Considering reconstruction metrics (CD, IoU) cannot capture generation diversity, we further evaluate the CLIP score of SOTA methods with texture generation on 100 in-the-wild objects. As illustrated in Table~\ref{tab:comparison_image_to_3d_texture}, the feed-forward paradigm of geometry and texture generation has achieved impressive performance and become the mainstream research direction of image-3D generation.

\begin{table}[t]
    \centering
    \scalebox{0.9}{
    \begin{tabular}{lc|lc}
        \hline
         Method & CLIP Score($\uparrow$) & Method & CLIP Score($\uparrow$) \\
         \hline
         TripoSG~\cite{li2025triposg} & 0.2243 & Hunayuan3D~\cite{yang2024hunyuan3d} & 0.2301 \\
         Meshy3D~\cite{Methy3d} & 0.2334 & Hitem3D~\cite{Hitem3d} & 0.2217 \\
         Trellis~\cite{xiang2025structured} & 0.2205 & Direct3D-S2~\cite{wu2025direct3ds2} & 0.2241 \\
         Seed3D~\cite{Methy3d} & 0.2310 & Step1x-3D~\cite{li2025step1x} & 0.2248\\
        \hline
    \end{tabular}}
    \caption{Quantitative comparison on image-to-3D methods with texture generation.}
    \label{tab:comparison_image_to_3d_texture}
\end{table}

\begin{table}[t!]
\footnotesize
\setlength{\tabcolsep}{4.5pt}
\centering

\begin{tabular}{clccc} 
\toprule
Type  &  Method  &  CD$\downarrow$ & IoU$\uparrow$ & Time$\downarrow$ \\
\midrule
\multirow{2}{*}{\begin{tabular}[c]{@{}c@{}}Optimization-based \\ Approaches\end{tabular}} & Realfusion~\cite{melas2023realfusion} & 0.0819 & 0.2741 & \textasciitilde 90min \\
& Magic123~\cite{qian2023magic123} & 0.0516 & 0.4528 & \textasciitilde 60min \\
\midrule
\multirow{3}{*}{\begin{tabular}[c]{@{}c@{}}MVS-based \\ Approaches\end{tabular}} & One-2-3-45~\cite{liu2024one} & 0.0629 & 0.4086 & \textasciitilde 45s \\
& Zero123~\cite{liu2023zero} & 0.0339 & 0.5035 & \textasciitilde 10min \\
& InstantMesh~\cite{xu2024instantmesh} & 0.0187 & 0.6353 & \textasciitilde 10s \\
\midrule
\multirow{4}{*}{\begin{tabular}[c]{@{}c@{}}Feedforward\\ Approaches\end{tabular}} & Point-E~\cite{nichol2022point} & 0.0426 & 0.2875 & \textasciitilde 40s \\
& Shap-E~\cite{shape-e} & 0.0436 & 0.3584 & \textasciitilde 10s \\
& Michelangelo~\cite{zhao2023michelangelo} & 0.0404 & 0.4002 & \textasciitilde 3s \\
& CraftsMan3D\cite{li2024craftsman} & 0.0291 & 0.5347 &\textasciitilde 5s \\
\bottomrule
\end{tabular}
\caption{ Quantitative comparison with baseline methods on the GSO dataset~\cite{downs2022google}. }
\label{tab:image-3d quantitative results}
\end{table}

\begin{figure}[t]
    \centering
    \includegraphics[width=1\linewidth]{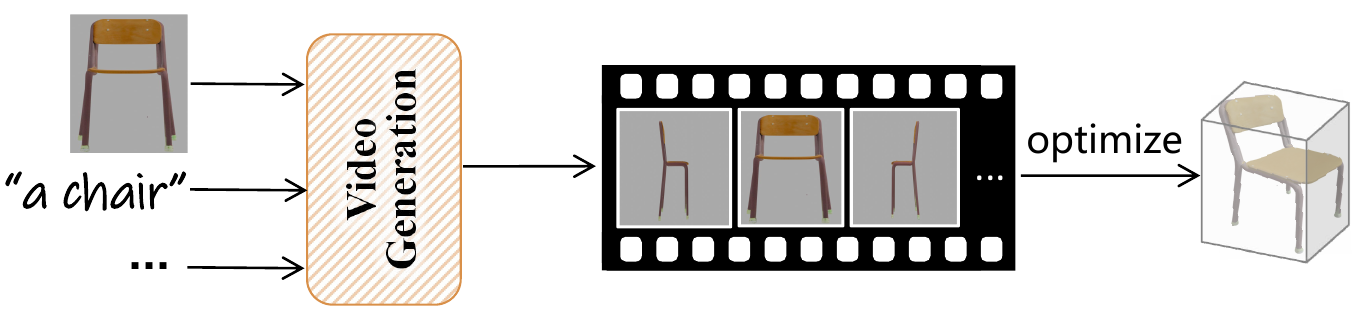}
    \caption{An illustration of video-to-3D generation paradigm. Camera-controllable video models can be used as a consistent multi-view generator for 3D generation.}
    \label{fig:relation_v23d}
\end{figure}

\noindent{\textbf{(3) Video-to-3D Generation.}} Video-to-3D methods build on 2D diffusion advances, extending image-based priors into time. By modeling coherent frame sequences with consistent texture, lighting, and geometry, video diffusion learns implicit 3D structure from motion and viewpoint changes. Frameworks like SV3D, Hi3D, and V3D adapt pretrained video diffusion backbones to generate multi-view frames, later fused via volume rendering, mesh optimization, or Gaussian splatting for explicit 3D reconstruction. This approach combines the synthesis strength of 2D models with spatial consistency and camera control crucial for dense 3D generation. The abundance of online video data further enriches 3D learning, offering temporal coherence and multi-view cues for realistic 3D scene synthesis~\cite{li2024textcraftor,wu2022object,videoworldsimulators2024}. As a result, leveraging these multi-view and time-variant data has become a promising approach for reconstructing and synthesizing 3D-consistent objects~\cite{sun2024t2v}. Recent research explores video-based priors for robust 3D generation ~\cite{gao2024cat3d,chen2024v3d,voleti2025sv3d}, aiming to learn 3D representations that maintain coherence across frames and adapt to changing viewpoints. At a high level, the main idea of these works for video-to-3D generation is to enable a camera-controllable video model as a consistent multi-view generator for dense 3D reconstruction (Fig.~\ref{fig:relation_v23d}).

Recent advancements in video diffusion models have illuminated their exceptional capabilities in generating realistic videos while providing implicit reasoning about 3D structures. However, significant challenges persist in employing these models for effective 3D generation, particularly regarding precise camera control. Traditional models~\cite{yu2023monohuman} are typically confined to generating clips with smooth and short camera trajectories, limiting their ability to create dynamic 3D scenes or integrate varying camera angles effectively. To address these limitations, several innovative techniques have been developed to enhance camera control within video diffusion frameworks. One early approach is AnimateDiff~\cite{guo2023animatediff}, employing Low-Rank Adaptation (LoRA)~\cite{hu2021lora} to fine-tune video diffusion models with fixed camera motion types, which facilitates the synthesis of structured scenes while adhering to specified camera dynamics. Another significant advancement is MotionCtrl~\cite{wang2024motionctrl}, which introduces conditioning mechanisms that allow models to follow arbitrary camera paths, thereby enabling greater flexibility in generating diverse perspectives and overcoming the rigidity of previous methods. Based on the availability of camera-controllable video generation, SVD-MV~\cite{blattmann2023stable}, SV3D~\cite{voleti2025sv3d}, and IM-3D~\cite{melas20243d} explore how to leverage camera control to improve the generation of 3D objects from video data. Specifically, SV3D trains a video diffusion model that renders arbitrary views, achieving strong generalization and high-resolution outputs (576$\times$576). This enables spatial coherence across frames and consistent rendering from multiple viewpoints, improving dense reconstruction. However, most such methods constrain camera motion to fixed orbital paths around central objects, limiting their ability to model complex scenes with diverse backgrounds and object interactions. As controlling camera movements in video models complements novel view information, several methods have explored the potential of video diffusion models for novel view synthesis (NVS). For example, Vivid-1-to-3~\cite{kwak2024vivid} effectively merges a view-conditioned diffusion model with a video diffusion model, allowing for the generation of temporally consistent views. By ensuring smooth transitions across frames, this model enhances the quality of the synthesized output, making it particularly effective for 3D scene representations. CAT3D~\cite{gao2024cat3d} augments rich multi-view information with a multi-view diffusion model.

\textbf{Discussion.} Leveraging video priors in multi-view generation transforms VDMs into consistent multi-view generators for dense 3D reconstruction. Further exploration will enhance high-fidelity 3D representation, especially for complex, dynamic environments requiring robust multi-view synthesis.

\subsubsection{Applications}
\label{sec:3D-apps}

\textbf{(1) Avatar Generation.} 
With the emergence of the metaverse and the popularity of VR/AR, 3D avatar generation has drawn increasing interest. Early works focus on generating head avatars \cite{zhang2023dreamface,han2024headsculpt}, which leverage text-to-image diffusion models and neural radiance fields to create facial assets. Later methods pay more attention to realistic full-body avatar generation by integrating neural radiance fields with statistical models \cite{kolotouros2024dreamhuman,huang2024dreamwaltz}. Recently, the animation capability of avatar generation has gained great attention with numerous methods being proposed \cite{tevet2022motionclip,tu2024motioneditor}. 
\textbf{(2) Scene Generation.} 
In addition to avatar generation, scene generation to create a realistic 3D environment is also highly demanded in applications like the metaverse and embodied intelligence. Early methods focus on object-centered scenes and leverage conditional diffusion models to synthesize multi-view images to optimize neural radiance fields \cite{nam20223d,li2023diffusion}. Later works extend these methods to room-scale scenes by introducing progressive strategies \cite{song2023roomdreamer,zhang2024text2nerf}. Motivated by their success, recent studies further investigate the generation of outdoor scenes, ranging from street-scale \cite{li2024dreamscene,lu2024urban} to city-scale \cite{zhou2025dreamscene360,deng2024citycraft}.
\textbf{(3) 3D Editing.}
The powerful 3D generation capability has created the downstream application of 3D content editing. Several methods focus on changing the appearance or geometry of the 3D content globally \cite{gao2023textdeformer,haque2023instruct} without isolating a specific region from a scene. For example, scene stylization methods \cite{fan2022unified,zhang2022arf} aim to manipulate the style of a 3D asset, such as illumination tuning and climate changes. Recent efforts have been made to achieve flexible 3D content editing at a finer-grained level. Specifically, appearance change \cite{sun2022fenerf,zhang2022fdnerf}, geometry deformation \cite{peng2022cagenerf,tseng2022cla}, and object-level manipulation \cite{kobayashi2022decomposing,wu2023objectsdf++} have been studied, with promising editing results being achieved.
\subsection{4D Generation}
\label{sec:4d}

We culminate in 4D generation by integrating all dimensions. As a cutting-edge field in computer vision, 4D generation focuses on synthesizing dynamic 3D scenes that evolve temporally based on multimodal inputs such as text, images, or videos. Unlike traditional 2D or 3D generation~\cite{tang2024edgerunner}, 4D synthesis introduces unique challenges, requiring both spatial coherence and temporal consistency while balancing high fidelity, computational efficiency, and dynamic realism~\cite{liu2024humangaussian,m2m}. In this section, we first introduce 4D representation, which builds upon 3D representation, and then summarize current 4D generation methods. Recent research has explored two primary paradigms: optimization-based methods leveraging SDS and feedforward-based approaches that avoid per-prompt optimization. These paradigms address distinct technical challenges, highlighting the field's complexity and the ongoing pursuit of a feasible balance among visual quality, computational efficiency, and scene flexibility. Representative works in 4D generation are summarized in Table~\ref{tab:4d_generation}.

\subsubsection{4D Representation}
\label{sec:4d-rep}

The field of 4D representation, which incorporates a temporal dimension into 3D modeling, provides a strong foundation for understanding dynamic scenes. By extending static 3D spatial representations $(x, y, z)$ with time $(t)$, these methods encode scene dynamics and transformations, essential for applications such as non-rigid human motion capture and simulating object trajectories~\cite{xian2021space, gao2021dynamic, li2022neural, li2021neural}. Most 4D representations can be decomposed into two components: a canonical 3D representation and a deformation model. The canonical 3D representation defines a static template shape, while the deformation model animates this template to generate motion. Common deformation representations include deformation fields—neural networks that map a space-time point to its corresponding position on the canonical shape—and deformation primitives, such as linear blend skinning (LBS), which express motion as a weighted combination of rigid transformations from different body parts or control points. Each approach has its trade-offs. Deformation fields are more flexible and can capture complex motions, whereas deformation primitives, tailored for articulated objects like humans or animals, offer greater robustness for large, structured motions. Deformation fields, lacking such inductive biases, often struggle with fast or large articulations. In the following discussion, we primarily focus on 4D representations based on canonical 3D templates combined with deformation fields.

A major challenge in 4D representation is the high computational cost of reconstructing a single scene. To address this, explicit and hybrid methods enhance efficiency without sacrificing quality. For instance, planar decompositions streamline 4D spacetime grids by breaking them into smaller components~\cite{cao2023hexplane, fridovich2023k, shao2023tensor4d}, while hash-based representations reduce memory and processing requirements~\cite{turki2023suds}. 3DGS balances speed and quality by using deformation networks to adapt static Gaussians to dynamic ones~\cite{kerbl20233d, luiten2024dynamic}. Recent advances disentangle static and dynamic scene components to render both rigid and non-rigid motions efficiently. For example, D-NeRF first encodes scenes into a canonical space and then maps them into temporally deformed states~\cite{pumarola2021d}. 3D Cinemagraphy generates feature-based point clouds from single images and animates using 3D scene flow~\cite{li20233d}. 4DGS captures temporal dynamics by modeling attributes like scales, positions, and rotations as time functions while keeping the static scene unchanged~\cite{wu20244d}. Hybrid NeRF-based methods expand 4D modeling with plane- and voxel-based feature grids. These grids, combined with MLPs, enable efficient novel-view synthesis and extend to dynamic scenes by incorporating temporal planes~\cite{cao2023hexplane, fridovich2023k}. Deformable NeRFs separate geometry and motion, simplifying motion learning and supporting applications like image-to-4D video generation and multi-view reconstruction~\cite{muller2022instant}. Collectively, these advancements reflect ongoing progress in achieving computationally efficient, high-quality temporal modeling for dynamic scenes.

\subsubsection{Algorithms}
\label{sec:4d-algorithms}

Modern 4D generation methods are largely rooted in the foundations laid by 3D generation. In particular, advances in 3D reconstruction, such as NeRF and 3DGS, have directly influenced how we model and render dynamic 4D scenes. These 3D frameworks provide not only efficient data structures and rendering techniques but also critical inductive biases that benefit temporal modeling in 4D. At the representation level, 3D methods offer canonical spatial priors that can be extended with deformation fields or motion trajectories to capture temporal evolution. From a training perspective, fast-training techniques from 3D (\textit{e.g.}, hash encoding, hierarchical sampling) have been adapted to accelerate 4D optimization. Human animation, as a most representative 4D task, particularly benefits from foundations and advancements in 3D human modeling. Techniques such as SMPL(-X), linear blend skinning, and neural deformation fields provide strong structural priors for modeling articulated motion. These 3D-based tools enhance both the realism and controllability of 4D human motion synthesis.

\begin{figure}
    \centering
    \includegraphics[width=0.9\linewidth]{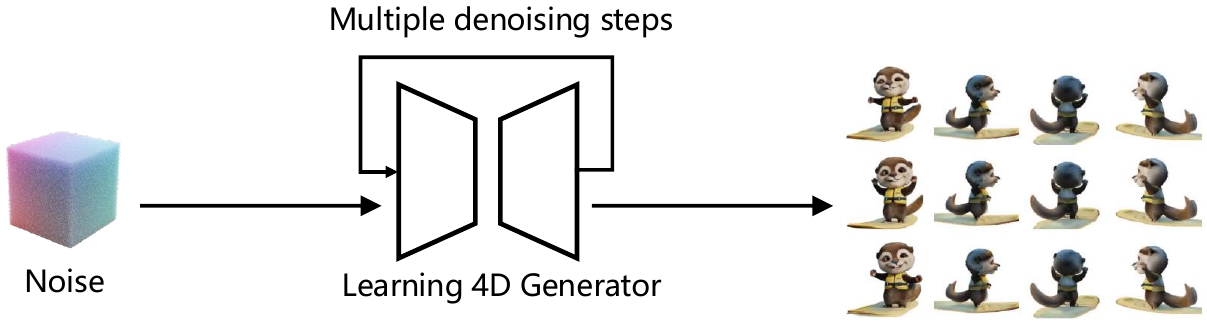}
    \caption{An illustration of the 4D generation feedforward paradigm, directly trained on the multi-view dynamic data.}
    \label{fig:4d_methods_ffw}
\end{figure}

\myPara{(1) Feedforward Approaches.} Feedforward-based methods provide an efficient alternative to generate 4D content in a single forward pass (Fig.~\ref{fig:4d_methods_ffw}), bypassing the iterative optimization required in SDS-based pipelines. These methods rely on pre-trained models, leveraging temporal and spatial priors to achieve fast and consistent generation. Control4D~\cite{shao2024control4d} and Animate3D~\cite{jiang2024animate3d} directly synthesize dynamic scenes from textual or visual inputs, enabling real-time applications such as interactive media. Vidu4D~\cite{wang2024vidu4d} improves motion trajectories by incorporating temporal priors, ensuring frame-to-frame coherence and smooth transitions. Diffusion4D~\cite{liang2024diffusion4d} extends the capabilities of diffusion models to handle 4D scene synthesis by combining spatial-temporal feature extraction with efficient inference mechanisms. L4GM~\cite{ren2024l4gm} further enhances feedforward techniques by integrating latent geometry modeling, producing high-quality results while maintaining computational efficiency.

\textbf{Discussion.} Feedforward-based approaches excel in scenarios that prioritize speed and adaptability. However, their reliance on pre-trained models and limited flexibility in handling complex dynamics pose challenges in achieving the same level of detail and diversity as optimization-based methods. Despite these limitations, feedforward techniques represent a significant step toward practical 4D generation, addressing key challenges of computational efficiency and scalability. By bridging the gap between quality and speed, these methods are poised to play a critical role in advancing 4D content generation for a wide range of applications.

\begin{figure}
    \centering
    \includegraphics[width=0.9\linewidth]{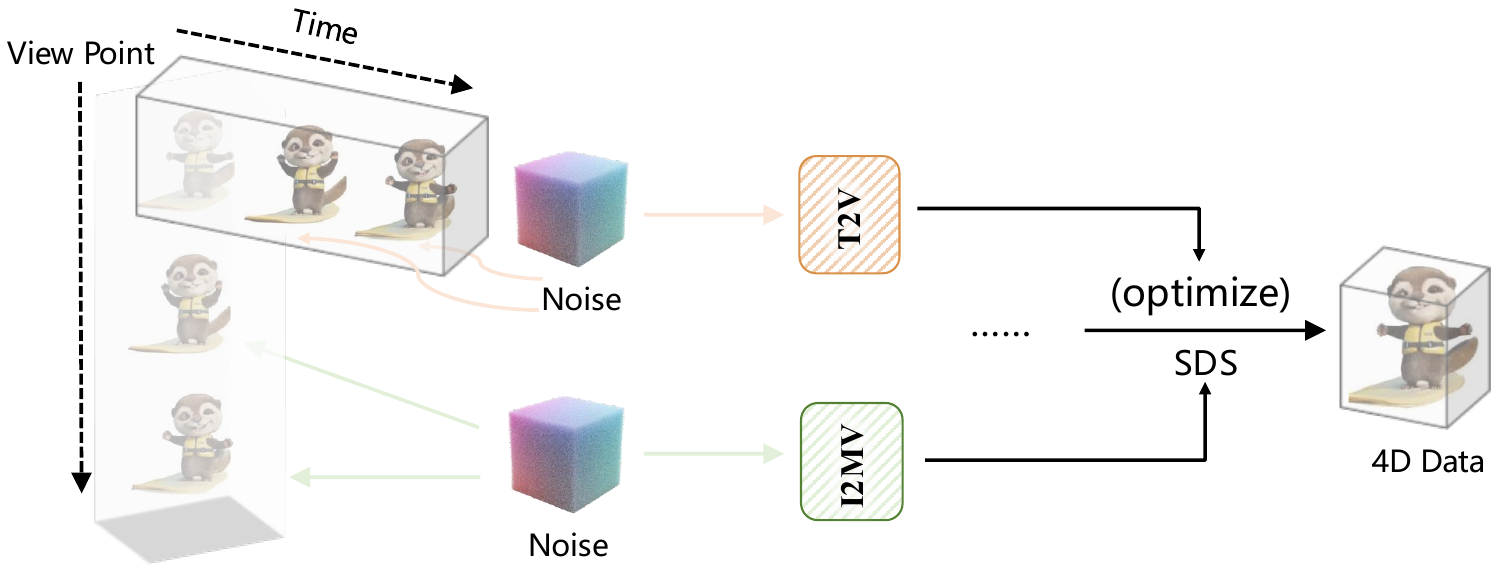}
    \caption{An illustration of 4D generation optimization-based paradigm. We use SDS guidance to make the optimized 4D asset follow the pretrained text-to-image, multi-view image, and text-to-video prior.}
    \label{fig:4d_methods_optim}
\end{figure}

\myPara{(2) Optimization-based Approaches.} Optimization-based methods are foundational in 4D generation, employing iterative techniques such as Score Distillation Sampling (SDS) to adapt pre-trained diffusion models to synthesize dynamic 4D scenes (Fig.~\ref{fig:4d_methods_optim}). These approaches leverage powerful priors from text-to-image, multi-view image, and text-to-video generation models, achieving temporally coherent scenes with rich motion dynamics. For example, MAV3D~\cite{singer2023text} optimizes NeRF or HexPlane features against SDS loss guided by textual prompts, while 4D-fy~\cite{bah20244dfy} and Dream-in-4D~\cite{zheng2024unified} improve the 3D consistency and motion dynamics by integrating image, multi-view, and video diffusion models in SDS supervision.
AYG~\cite{ling2024align} proposes to use deformable 3DGS as an inherent representation, easily disentangling static geometry from motion with a simple delta deformation field to improve flexibility. 

Based on such pipelines, recent works further improve 4D generation from multiple aspects: appearance quality, geometry consistency, motion faithfulness, and generation controllability. In particular, TC4D~\cite{bahmani2025tc4d} and SC4D~\cite{wu2025sc4d} enable free user control on 4D object motion trajectory. STAG4D~\cite{zeng2025stag4d} employs multi-view fusion to enhance spatial and temporal alignment across frames, ensuring smooth transitions and consistency. Additionally, DreamScene4D~\cite{chu2024dreamscene4d} and DreamMesh4D~\cite{li2024dreammesh4d} adopt disentanglement strategies to localize optimization efforts, significantly reducing computational overhead while maintaining high fidelity. Recent advances such as 4Real~\cite{yu20244real} and C3V~\cite{zhu2024compositional} further push the boundaries of optimization-based methods by combining compositional scene generation with efficient optimization. These methods break dynamic scenes into modular components such as static geometry and motion fields, enabling flexible updates and diverse content generation. Despite their strengths in achieving high-quality and temporally consistent results, optimization-based approaches remain computationally demanding, with runtime requirements often precluding real-time applications. As research progresses, ongoing efforts focus on improving scalability and reducing latency without compromising visual fidelity or dynamic realism.

\begin{figure}[t]
    \centering
    \includegraphics[width=0.9\linewidth]{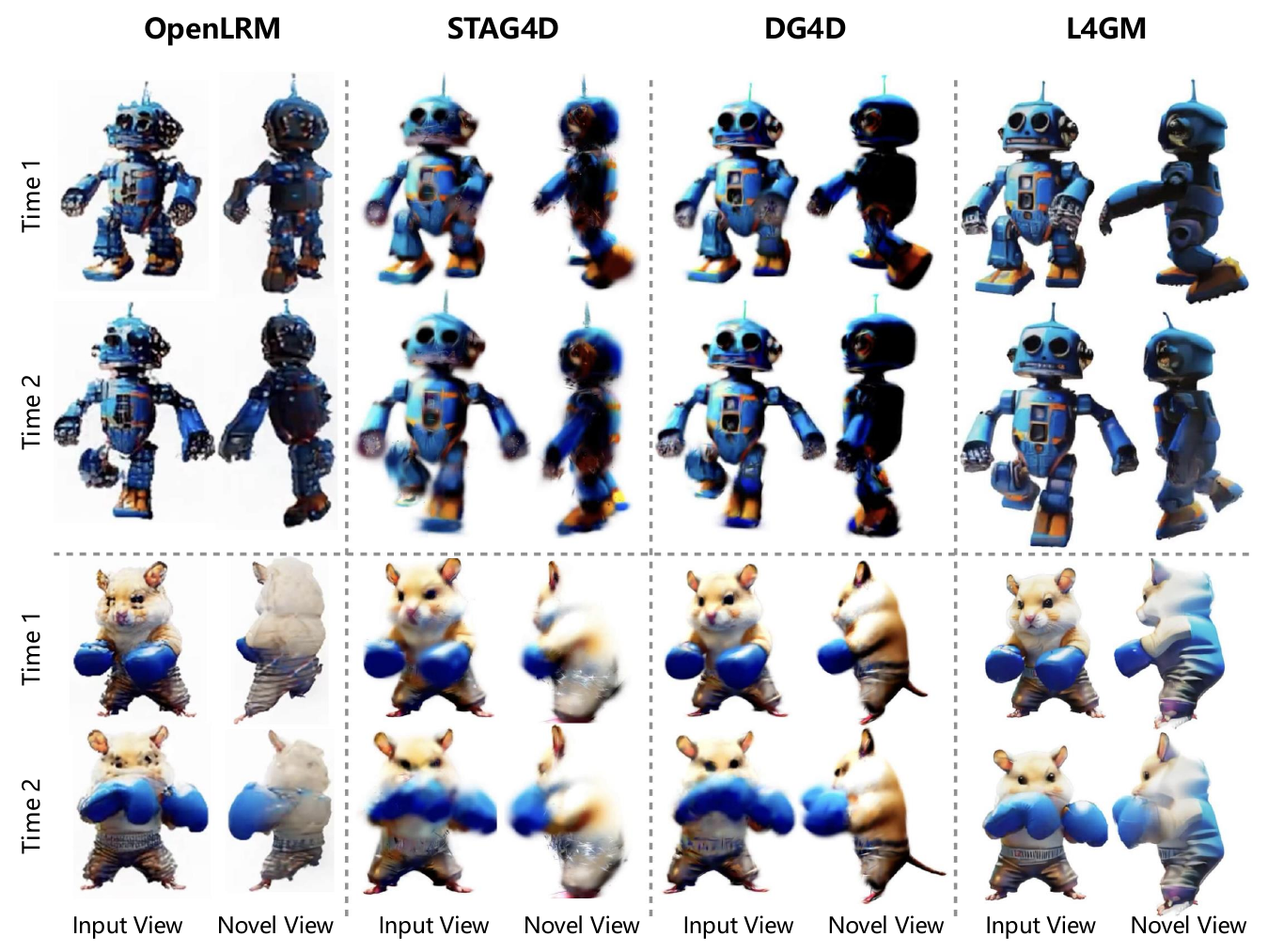}
    \caption{Qualitative comparison between different 4D generation methods. Results are obtained from L4GM~\cite{ren2024l4gm}.}
    \label{fig:comparison_4d}
\end{figure}

\begin{table}[t]
    \centering
    \scalebox{1.0}{
    \begin{tabular}{l|c|c|c}
        \hline
         & \textbf{Time ($\downarrow$)} & \textbf{CLIP ($\uparrow$)} & \textbf{LPIPS ($\downarrow$)}  \\
         \hline
         Consistent4D \cite{jiang2023consistent4d} & 2 hr & 0.87 & 0.16 \\ 
         4DGen \cite{yin20234dgen} & 1 hr & 0.89 & 0.13 \\
         STAG4D \cite{zeng2025stag4d} & 1 hr & 0.91 & 0.13 \\ 
         DG4D \cite{ren2023dreamgaussian4d} & 10 min & 0.87 & 0.16 \\
         \hline
    \end{tabular}}
    \caption{Quantitative comparison between different 4D generation methods. Results are obtained from \cite{ren2024l4gm}.}
    \label{tab:comparison_4d}
    \vspace{-0.3cm}
\end{table}

\textbf{Evaluation.} Quantitative evaluation of 4D generation methods mainly focuses on three aspects: 1) Time cost to generate a single 4D asset. 2) CLIP score to evaluate image similarity between the generated and the ground truth novel views. 3) Perceptual similarity (LPIPS) between the generated and the ground truth novel views. We present quantitative results in Table~\ref{tab:comparison_4d} and visual results in Fig.~\ref{fig:comparison_4d}.

Feedforward 4D generators amortize computation, achieving seconds-to-minutes asset generation and often competitive CLIP and FVD scores under moderate camera motion. However, their performance can degrade on large-baseline novel views and occlusions due to the implicit representation of scene states. In contrast, SDS and other optimization-based pipelines explicitly reconstruct scenes, offering improved multi-view consistency and editability, but at the cost of significantly higher computational time (minutes to hours) and memory usage. Representation choice accounts for much of the variance: Gaussians offer speed and sharpness but can flicker under fast motion; plane-factorizations (e.g., K-/HexPlane) maintain stability across wide baselines; canonical-space with deformation fields provides robustness for non-rigid dynamics.
Supervision also plays a key role: text-only SDS is under-constrained, whereas multiview or video guidance strengthens geometric and temporal coherence. Think of this as a Pareto frontier: feedforward methods favor speed and adaptability, SDS pipelines maximize fidelity and editability, and compositional or controllable variants trade some frame quality for stronger constraints and reusability.

\subsubsection{Applications}
\label{sec:4D-apps}

\textbf{(1) 4D Editing.} Instruction-guided editing, which allows users to edit scenes through natural language, offers a user-friendly and intuitive approach. While this has been successfully achieved for 2D images with models like Instruct-Pix2Pix (IP2P)~\cite{brooks2023instructpix2pix} and Instruct-NeRF2NeRF (IN2N)~\cite{haque2023instruct} for 3D scenes, extending this capability to 4D scenes presents significant challenges. Recent developments in text-to-image diffusion models and differentiable scene representations have enabled editing 4D scenes using text prompts. For instance, Instruct 4D-to-4D~\cite{instruct-4d-to-4d} treats 4D scenes as pseudo-3D scenes, adopting a video editing approach to iteratively generate coherent edited datasets. Concurrent work like Control4D~\cite{shao2024control4d} employs GANs alongside diffusion models to ensure consistent editing for dynamic 4D portraits based on text instructions. 
\textbf{(2) Human Animation.}  
As a core element of 4D simulation, human motion generation has become a central focus of research in this field. Different from the human-centric video generation (in Sec.~\ref{sec:Video-apps}), 3D human motion generation makes it easier to animate characters in 3D applications, \textit{e.g.} games and embodied intelligence. The recent success of the 3D human motion generation mainly comes from the well-studied human body parameterized models~\cite{smpl,smplx}. The human motion generation aims to achieve a basic target: simulating 4D human subjects in the digital world, which can be divided into two folds. 1)~\textit{Generating motions with sparse control signals} mainly simulates human animations in the virtual world based on the sparse motions specified by users. Robust motion in-between~\cite{robustmotion} proposes a time-to-arrival embedding and the scheduled target noise vector to perform motion in-between with different transition lengths in a model robustly. As the phase manifold of the motion space has a good structure, Starke \textit{et al.}~\cite{phasebetween} propose a mixture-of-expert network to interpolate between motions in the phase manifold. Besides, another part of sparse control-guided motion generation is motion prediction, \textit{a.k.a} motion extrapolation. Early works~\cite{martinez2017human, corona2020context} model motion prediction deterministically, while later studies~\cite{mtvae, dlow, humanmac,feng2024motionwavelet} introduce stochastic approaches to capture its inherent subjectivity.
2)~\textit{Generating motions with multi-modality conditions} aims to simulate human motions with other modality inputs, \textit{e.g.} text, audio, musics. To resolve the problem of limited pairwise text-motion data, Guo \textit{et al.}~\cite{guo2022generating} propose a relatively large text-motion dataset, HumanML3D, which is significantly larger than previous datasets and greatly pushes the development of the tasks. VQ-VAE~\cite{humantomato,motiongpt,momask,scamo,motionclr} effectively supports text-to-motion synthesis, while diffusion models~\cite{mdm,mofusion,motionlcm} further improve generation quality. Music-driven dance generation follows similar paradigms—cVAE~\cite{danceformer}, VQ-VAE~\cite{bailando,chen2025motionllm}, and diffusion-based~\cite{lodge} methods.

\section{Future Directions}
\label{sec:direction}

Despite advances in 2D, video, 3D, and 4D generation, many open questions remain about how these modalities could interact. We propose viewing them as a coupled dimensional hierarchy: lower-dimensional models provide priors for higher-dimensional synthesis, while higher-dimensional structure regularizes lower-dimensional generation. Under this view, the main bottleneck of 4D generation is not only limited to 4D data, but also the lack of principled ways to transfer rich semantics from large-scale 2D corpora into coherent spatiotemporal representations.

\noindent\textbf{From Isolation to Synergy: A Coupled Dimensional Hierarchy.}
Our survey framework suggests that progress across dimensions is fundamentally coupled rather than parallel. In particular, 3D and video generation can be interpreted as two different ``derivatives'' of 2D priors~\cite{poole2024dreamfusion, rombach2022high}: 3D focuses on enforcing multi-view geometric consistency~\cite{liu2023zero, liu2024syncdreamer}, while video focuses on enforcing temporal continuity~\cite{guo2023animatediff}. 4D generation further requires these two forms of consistency to hold simultaneously~\cite{jiang2023consistent4d, singer2023text}. This hierarchy implies that improvements in 2D realism and semantic alignment often determine the fidelity ceiling of downstream 3D/4D synthesis. Therefore, future research should emphasize cross-dimensional synergy, where representation learning, supervision signals, and constraints are explicitly designed to transfer knowledge across dimensions.

\noindent\textbf{Semantic Offloading to 2D Foundation Models (2D/Video $\to$ 3D/4D).}
A fundamental asymmetry exists in data scale and diversity: 2D image-text pairs are orders of magnitude more abundant than high-quality 3D/4D supervision. For example, while some datasets like ~\cite{deitke2023objaverse, deitke2024objaverse} provide large-scale 3D data, they remain tiny compared to massive 2D datasets such as LAION-5B~\cite{schuhmann2022laion}. This suggests that 2D foundation models should serve as the semantic and diversity engine for higher-dimensional generation. In particular, many challenging tasks such as fine-grained editing, compositional control, and semantic diversity should be offloaded to the 2D domain, where models are already highly capable and training signals are abundant. Higher-dimensional models can then focus on the lifting process~\cite{lin2023magic3d, wang2024prolificdreamer}, namely reconstructing consistent geometry and motion from these 2D priors. From this viewpoint, scalable 4D generation depends not only on collecting more 4D data, but also on developing efficient projection operators that can map rich 2D distributions into 3D/4D representations while preserving controllability and diversity.

\noindent\textbf{Consistency Back-propagation and Physics-grounded Regularization  (3D/4D $\to$ 2D/Video).}
The interaction across dimensions should not be viewed as a one-way pipeline. High-dimensional structural constraints can back-propagate to improve low-dimensional generation. For example, 3D/4D synthesis inherently requires coherent geometry, correspondence, and identity preservation across both views and time. These constraints can serve as strong regularizers for current video diffusion models, which still suffer from long-term temporal inconsistency, such as flickering, drifting appearance, and unstable motion trajectories. More broadly, physics-grounded constraints (e.g., collision avoidance, material consistency, and plausible dynamics) provide a form of supervision that is difficult to obtain in pure 2D generation. This suggests that 4D generation is not only an application of 2D priors, but also a potential mechanism for enforcing physically meaningful consistency that may address long-standing weaknesses in low-dimensional generative models.

\noindent\textbf{Toward Unified Spatiotemporal World Models.}
A natural consequence of the above insights is that future generative systems may require a unified backbone that jointly models spatial reconstruction (2D $\to$ 3D) and temporal evolution (2D $\to$ video). Existing pipelines often treat 3D generation and video generation as separate extensions of 2D diffusion, but both can be seen as complementary constraints on the same underlying world state. A unified architecture that simultaneously reasons about spatial structure and temporal dynamics could enable consistent 4D synthesis with stronger controllability. Such a backbone may involve shared latent representations, explicit correspondence modeling, and joint conditioning mechanisms that bridge view synthesis and motion prediction. Ultimately, this direction points toward generative world models~\cite{agarwal2025cosmos} that unify appearance, geometry, and dynamics into a single spatiotemporal representation.

\section{Conclusions}
\label{sec:conclusions} 
In this survey, we have reviewed recent advances in multimodal generative models for simulating the real world, focusing on the intertwined dimensions of appearance, dynamics, and geometry. We categorized existing approaches across 2D, video, 3D, and 4D generation, and discussed their representative methodologies, cross-domain relations, and technical differences, supported by comparative visual examples. Furthermore, we summarized commonly used datasets and evaluation metrics to provide a practical reference for benchmarking.

Despite the rapid progress, fundamental challenges persist, such as the scalability of generative pipelines, temporal consistency in long-range sequences, and adaptability to real-world dynamics. We identified several open research directions, including the need for unified representations across modalities, efficient training with sparse supervision, and the integration of physics-based constraints for enhanced realism.

We hope this survey provides not only a comprehensive overview for newcomers but also a foundation for future research towards more coherent, controllable, and physically grounded multimodal generation systems.

{\small
\bibliographystyle{IEEEtran}

\bibliography{egbib}
}

\newpage

\ifCLASSOPTIONcaptionsoff
  \newpage
\fi

\newpage
\appendices
\section{Glossary of Technical Terms}
\label{sec:glossary}

This glossary provides definitions and references for key technical terms and acronyms used throughout this survey to assist readers in understanding domain-specific concepts.

\subsection{Neural Representations}

\noindent\textbf{NeRF (Neural Radiance Field)~\cite{mildenhall2020nerf}}: A continuous volumetric function encoded in a Multi-Layer Perceptron (MLP) that maps 3D spatial positions and viewing directions to density and color values, enabling photorealistic novel view synthesis through volumetric rendering.

\noindent\textbf{3DGS (3D Gaussian Splatting)~\cite{kerbl20233d}}: An efficient 3D scene representation that models objects as collections of anisotropic Gaussian distributions with learnable parameters (position, covariance, opacity, and appearance), enabling fast training and real-time rendering.

\noindent\textbf{SDF (Signed Distance Field/Function)~\cite{park2019deepsdf}}: An implicit 3D representation that encodes geometry by storing the signed distance from any point in space to the nearest surface, with negative values inside objects and positive values outside.

\noindent\textbf{Triplane~\cite{chan2022efficient}}: A memory-efficient 3D representation that decomposes 3D volumes into three orthogonal 2D feature planes (XY, XZ, YZ), enabling faster rendering and reduced memory consumption compared to voxel-based approaches.

\noindent\textbf{DMTet (Deep Marching Tetrahedra)~\cite{shen2021dmtet}}: A hybrid 3D representation that combines implicit SDF-based modeling with explicit tetrahedral mesh extraction, enabling differentiable mesh generation with topology flexibility.

\subsection{Generative Models}

\noindent\textbf{Diffusion Models~\cite{ho2020denoising,rombach2022high}}: A class of generative models that learn to generate data by iteratively denoising samples from a Gaussian noise distribution, guided by learned score functions.

\noindent\textbf{VAE (Variational Autoencoder)~\cite{kingma2013auto}}: A generative model that learns a probabilistic mapping between input data and a latent representation, enabling generation by sampling from the learned latent space.

\noindent\textbf{VQ-VAE (Vector Quantized VAE)~\cite{van2017neural}}: An extension of VAE that uses discrete latent representations through vector quantization, enabling more stable training and better reconstruction quality.

\noindent\textbf{GAN (Generative Adversarial Network)~\cite{goodfellow2020generative}}: A generative model framework consisting of a generator and discriminator network trained adversarially to generate realistic samples.

\noindent\textbf{LDM (Latent Diffusion Model)~\cite{rombach2022high}}: Also known as Stable Diffusion, a diffusion model that operates in a compressed latent space rather than pixel space, significantly improving efficiency while maintaining high generation quality.

\subsection{Training and Optimization}

\noindent\textbf{SDS (Score Distillation Sampling)~\cite{poole2024dreamfusion}}: A distillation technique introduced in DreamFusion that enables optimization of 3D representations using pre-trained 2D diffusion models as supervision, transferring their learned priors to 3D generation without requiring 3D training data.

\noindent\textbf{CLIP (Contrastive Language-Image Pre-training)~\cite{radford2021learning}}: A vision-language model trained on image-text pairs that learns aligned embeddings for images and text, enabling zero-shot image classification and serving as a powerful prior for multimodal generation tasks.

\subsection{3D Generation Paradigms}

\noindent\textbf{MVS (Multi-View Stereo)}: A 3D reconstruction technique that synthesizes 3D geometry from multiple 2D images captured from different viewpoints. In the context of 3D generation, MVS-based approaches first generate multi-view images and then reconstruct 3D models from them.

\noindent\textbf{Optimization-based Approaches}: Methods that iteratively optimize 3D representations (e.g., NeRF parameters) for each input prompt using techniques like SDS loss, achieving high quality at the cost of longer generation times.

\subsection{Key Systems and Frameworks}

\noindent\textbf{DreamFusion~\cite{poole2024dreamfusion}}: A pioneering text-to-3D generation system that introduced Score Distillation Sampling (SDS) to optimize NeRF representations using pre-trained text-to-image diffusion models.

\noindent\textbf{Stable Diffusion (SD)~\cite{rombach2022high}}: A widely-used open-source latent diffusion model for high-quality T2I generation, serving as a foundation for many multimodal generation systems.

\noindent\textbf{Instant-NGP~\cite{muller2022instant}}: An efficient neural graphics primitive framework using multi-resolution hash encoding for fast NeRF training and rendering.

\noindent\textbf{D-NeRF~\cite{pumarola2021d}}: Dynamic Neural Radiance Field that extends NeRF to model temporal changes by mapping scenes from a canonical space to time-varying deformed states.

\noindent\textbf{4DGS (4D Gaussian Splatting)~\cite{wu20244d}}: An extension of 3D Gaussian Splatting that models dynamic scenes by representing Gaussian attributes (position, scale, rotation) as time-dependent functions.

\subsection{Additional Terminology in This Paper}

\noindent\textbf{Cross-attention}: An attention mechanism that allows a model to attend to features from different modalities or sources, commonly used to inject conditioning information (e.g., text embeddings) into generation models.

\noindent\textbf{Volume Rendering}: A technique for generating 2D images from 3D volumetric data by integrating color and density along viewing rays, central to NeRF-based rendering.

\noindent\textbf{Deformation Field}: A function that maps points from a canonical 3D space to deformed positions, enabling modeling of non-rigid motion and dynamic scenes.

\noindent\textbf{Canonical Space}: A reference coordinate system where objects are represented in a standard pose or configuration, facilitating learning of deformations and articulations.

\noindent\textbf{Spherical Harmonics}: A mathematical basis for representing functions on a sphere, used in 3DGS to encode view-dependent appearance efficiently.

\noindent\textbf{Hash Encoding~\cite{muller2022instant}}: A technique using learnable hash tables to encode spatial features at multiple resolutions, enabling fast training and inference in neural rendering.

\subsection{Note on Abbreviations}

Throughout this survey, we use standard abbreviations for common model architectures and techniques:
\begin{itemize}
    \item \textbf{2D/3D/4D}: Referring to two-dimensional (images), three-dimensional (spatial geometry), and four-dimensional (spatial + temporal) content.
    \item \textbf{T2I/T2V/T23D/T24D}: Text-to-Image/Video/3D/4D generation.
    \item \textbf{I23D/V23D}: Image-to-3D/Video-to-3D generation.
    \item \textbf{DiT}: Diffusion Transformer. The seminal paper refers to ~\cite{peebles2023scalable}. 
    \item \textbf{U-Net}: A convolutional neural network architecture with encoder-decoder structure and skip connections. 
\end{itemize}

\noindent For comprehensive technical details and mathematical formulations of these concepts, readers are encouraged to consult the cited references and Section~\ref{sec:background} (Background and Preliminaries) of this survey.
\section{Preliminaries} 
\label{sec:background}

Deep generative models learn complicated and high-dimensional data distributions with the aid of deep neural networks. Denote the data sample as $\mathbf{x}$ and its  distribution as $p_{data}(\mathbf{x})$, the objective of deep generative models is to approximate the $p_{data}(\mathbf{x})$ with $p_{\theta}(\mathbf{x})$, where $\theta$ is the parameter of models. In this section, we briefly review several mainstream generative models (Table~\ref{tab:pre}), including generative adversarial networks (GANs)~\cite{goodfellow2020generative}, variational autoencoders (VAEs)~\cite{kingma2013auto}, autoregressive models (AR Models)~\cite{bengio2000neural}, normalizing flows (NFs)~\cite{rezende2015variational}, and diffusion models~\cite{ho2020denoising}.

\subsection{Generative Adversarial Networks (GANs)}

GANs avoid the parametric form of $p_{\theta}(\mathbf{x})$ but represent $p_{\theta}(\mathbf{x})$ as the distribution of samples produced by a generator. It has been shown that $p_{\theta}(\mathbf{x})$ will converge to $p_{data}(\mathbf{x})$ under certain conditions~\cite{goodfellow2020generative}. 

Specifically, the generator takes a noise vector $\mathbf{z}$ as input to synthesize a data sample $G(\mathbf{z};\theta_g)$. The $p_{\theta}(\mathbf{x})$ is defined as the distribution of $G(\mathbf{z};\theta_g)$, where $\mathbf{z}\sim p_{\mathbf{z}}(\mathbf{z})$. Meanwhile, the discriminator $D(\mathbf{x}; \theta_d)$ identifies whether an input data sample is a real one or a synthetic one. During training, the discriminator is trained to distinguish the generated samples from real data, while the generator is trained to deceive the discriminator. The process can be formulated as:
\begin{equation}
\begin{aligned}
    \min_G \max_D V(G,D) = \mathbb{E}_{\mathbf{x}\sim p_{data}(\mathbf{x})}[\log D(\mathbf{x};\theta_{d})]&+\\
    \mathbb{E}_{\mathbf{z}\sim p_{\mathbf{z}}(\mathbf{z})}[\log (1-D(G(\mathbf{z};\theta_g);\theta_d))]&.
\end{aligned}
\end{equation}
There are several challenges associated with GAN training. For example, the Nash equilibrium may not always exist~\cite{farnia20a} or be hard to achieve~\cite{jabbar2021survey}, resulting in unstable training. Another problem is mode collapse, where the generator produces only specific types of samples with low diversity~\cite{jabbar2021survey,arora2017generalization}.

\subsection{Variational Autoencoders (VAEs)}
The variational autoencoders (VAEs) formulate the $p_{\theta}(\mathbf{x})$ as,
\begin{equation}
   p_{\theta}(\mathbf{x})=\int p_{\theta}(\mathbf{x}|\mathbf{z})p_{\theta}(\mathbf{z}) d\mathbf{z}, 
\end{equation}
where $p_{\theta}(\mathbf{z})$ is a prior distribution of $\mathbf{z}$, and $p_{\theta}(\mathbf{x}|\mathbf{z})$ is the distribution of $\mathbf{x}$ conditioned on $\mathbf{z}$. However, since this integration is usually intractable, VAEs maximize the lower bound of $\log p_{\theta}(\mathbf{x})\geq -\mathtt{KL}(q_{\theta}(\mathbf{z}|\mathbf{x})||p_{\theta}(\mathbf{z}))+\mathbb{E}_{q_{\theta}(\mathbf{z}|\mathbf{x})}[\log p_{\theta}(\mathbf{x}|\mathbf{z})]$, where $\mathtt{KL}(q_{\theta}(\mathbf{z}|\mathbf{x})||p_{\theta}(\mathbf{z}))$ is the KL divergence between $q_{\theta}(\mathbf{z}|\mathbf{x})$ and $p_{\theta}(\mathbf{z})$, and $\mathbb{E}_{q_{\theta}(\mathbf{z}|\mathbf{x})}[\log p_{\theta}(\mathbf{x}|\mathbf{z})]$ is computed by the Stochastic Gradient Variational Bayes estimator~\cite{kingma2013auto}. The challenges for VAEs include undesirable stable equilibrium, blurriness, and so on~\cite{kingma19int}.

\subsection{Autoregressive Models (AR Models)}
Autoregressive (AR) models factorize $p_{\theta}(\mathbf{x})$ as a product of conditional probabilities,
\begin{equation}
    p_{\theta}(\mathbf{x}) = p(x_1,\ldots,x_{d}) = \prod_{i=1}^{d} p_{\theta}(x_i \mid x_1,\ldots,x_{i-1}),
\end{equation}
where $d$ is the sequence length.  
Such factorization simplifies multivariate density estimation and has been widely adopted to model pixels sequentially in images~\cite{van2016conditional,jain2020lmconv,hoogeboom2022autoregressive}.

To reduce the quadratic cost of attention in standard transformer-based AR models, several non-transformer architectures have recently been introduced. RWKV~\cite{peng2023rwkv}, Mamba~\cite{gu2024mamba}, and RetNet~\cite{sun2023retentive} replace or augment attention with recurrent or state-space mechanisms. RWKV and Mamba employ purely recurrent designs that maintain a fixed-size memory, delivering linear-time inference over moderate sequence lengths but still facing challenges at extreme context sizes. RetNet updates hidden states through a retention mechanism, providing an efficient alternative to global self-attention. Although these architectures show promising results on language and other sequence tasks, their use as backbones for deep generative models is still limited. Future work that integrates them into generative pipelines may improve the trade-off among sample quality, scalability, and memory usage.

\begin{table}[t]
    \centering
    \caption{Comparison of deep generative models.}
    \resizebox{\linewidth}{!}{
    \begin{tabular}{lll}
    \toprule
    \textbf{Model Types} & \textbf{Advantages} & \textbf{Disadvantages} \\ 
    \midrule
    {GANs~\cite{goodfellow2020generative}} 
        & 1) Flexible 
        & 1) Unstable training \\ 
         & 2) Efficient inference & 2) Lack diversity \\
         & & 3) Mode collapse\\
    \midrule
    {VAEs~\cite{kingma2013auto}} 
        & 1) Data compression & 1) Posterior collapse\\
        &  & 2) Blur \\ 
    \midrule
    \multirow{2}{*}{AR Models~\cite{bengio2000neural}} 
        & 1) Explicit density  & 1) Markov assumption  \\
        &                   & 2) Difficult to parallelize \\ 
    \midrule
    {NFs~\cite{rezende2015variational}} 
        & 1) Explicit density  & 1) Limited capacity \\ 
        & & 2) Restricted architecture \\
    \midrule
    {Diffusion Models~\cite{ho2020denoising}} 
        & 1) High-quality samples & 1) Expensive computation \\
        & 2) Learn complex distribution & \\
    \bottomrule
    \end{tabular}
    }
    \label{tab:pre}
\end{table}

\subsection{Normalizing Flows (NFs)} 
NFs employ an invertible neural network $g(\cdot)$ to map $\mathbf{z}$ from a known and tractable distribution to the real data distribution. In this way, $p_{\theta}(\mathbf{x})$ can be formulated as,

\begin{equation}
    p_{\theta}(\mathbf{x})=p(f(\mathbf{x}))|\mathtt{det} \ \mathtt{J}(g(g^{-1}(\mathbf{x})))|^{-1},
\end{equation}
where $g^{-1}(\cdot)$ is the inverse of $g(\cdot)$, and $\mathtt{J}(g(\mathbf{z}))=\frac{\partial g}{\partial \mathbf{z}}$ is the Jacobian of $\mathbf{g}$. NFs construct more complicated non-linear invertible functions by compositing a set of $N$ bijective functions and define $g=g_N\circ\dots\circ g_1$. 

\subsection{Diffusion Models} 
Diffusion models are a class of probabilistic generative models that iteratively corrupt data by introducing noise and subsequently learn to reverse this process to generate samples. We define $p_{\theta}(\mathbf{x})$ with an energy term $s_{\theta}(\mathbf{x})$,
\begin{equation}
    p_{\theta}(\mathbf{x})={\exp(-s_{\theta}(\mathbf{x}))}/{Z_{\theta}},
\end{equation}
where $Z_{\theta}=\int \exp(-s_{\theta}(\mathbf{x})) d\mathbf{x}$ is the normalization term. Since $Z_{\theta}$ is intractable to compute, the diffusion models learn the score function $\nabla_{\mathbf{x}}\log p_{\theta}(\mathbf{x})=-\nabla_{\mathbf{x}}s_{\theta}(\mathbf{x})$ instead.

The forward process that transforms data distribution to standard Gaussian is defined by $d\mathbf{x} = f(\mathbf{x}, t)dt + g(t)d\mathbf{w}$, where $f(\mathbf{x}, t)$ is the drift coefficient, $g(t)$ is the diffusion coefficient,  $\mathbf{w}$ is the standard Wiener process, and $t\in[0,1]$. The samples are generated by the corresponding reverse process, which is described by $d\mathbf{x}=[f(\mathbf{x}, t)-g^2(t)\nabla\log p_t(\mathbf{x})]dt+g(t)d\mathbf{\overline{w}}$, where $\mathbf{\overline{w}}$ is a standard Wiener process when time flows backward from 1 to 0.

\section{Datasets and Evaluations}
\label{sec:eval}

\begin{table*}[!t]
    \center
    \caption{Summary of the widely-used 2D, video, 3D and 4D generation datasets. \textcolor{magenta}{[Link]} directs to dataset websites.}
    \scalebox{0.54}{
    \begin{threeparttable}
    \setlength{\tabcolsep}{1mm}{
    \begin{tabular}{llcccllp{0.4\textwidth}p{0.4\textwidth}} 
        \toprule[1pt]
        \textbf{Dataset} & \textbf{Year} & \textbf{\# Categories} & \textbf{\# Objects} & \textbf{\# Images} & \textbf{Type} &\textbf{Evaluation Metrics} & \textbf{Contribution Highlights} & \textbf{Limitations} \\
        \midrule

        \belowrulesepcolor{gray!15!}
        \multicolumn{9}{c}{\textbf{2D Generation (Sec.~\ref{sec:2D})}} \\ \aboverulesepcolor{gray!15!} \midrule

        SBU~\cite{ordonez2011im2text}~\href{https://www.cs.rice.edu/~vo9/sbucaptions/}{[Link]} &2011 & 89 & - & 1M & Image-Text pairs & - & Photographs with captions and 5 kinds of image content. & Noisy user captions and single reference.\\
        MS-COCO~\cite{lin2014microsoft}~\href{https://cocodataset.org/}{[Link]} &2014 & 80 for objects, 91 for stuffs & 1.5M & 330K & Image-Text pairs & FID, CLIP-Score, SSIM & Manual annotation and 330K for English. & Class imbalance and social bias.\\
        CC-3M~\cite{sharma2018conceptual}~\href{https://ai.google.com/research/ConceptualCaptions/}{[Link]} &2018 & - & - & 3M & Image-Text pairs & - & Alt-text caption annotation. & Noisy user captions and weakly aligned.\\
        LAION-5B~\cite{schuhmann2022laion}~\href{https://laion.ai/blog/laion-5b/}{[Link]} &2022 & - & - & 5.85B & Image-Text pairs & FID, CLIP-Score, SSIM &  5.85B CLIP-filtered multilingual image-text pairs. & Uncurated and unsafe contents.\\

        ShareGPT4V ~\cite{chen2025sharegpt4v}~\href{https://huggingface.co/datasets/Lin-Chen/ShareGPT4V}{[Link]} &2023 & - & - & 102K & Image-Text pairs & - &  1.2M image-text pairs for ShareGPT4V-PT. & AI generated captions with hallucination.\\ 
        
        Pick-a-Pic~\cite{kirstain2023pick}~\href{https://github.com/yuvalkirstain/PickScore}{[Link]} &2023 & - & - & 500K & Image-Text pairs & FID, PickScore &  Human preferences on model-generated images. & Limited preference labels.\\
        \midrule

         \belowrulesepcolor{gray!15!}
        \multicolumn{9}{c}{\textbf{Video Generation (Sec.~\ref{sec:video})}} \\ \aboverulesepcolor{gray!15!} \midrule

        UCF-101~\cite{soomro2012ucf101}~\href{https://www.crcv.ucf.edu/data/UCF101.php}{[Link]} & 2012 & 101 & - & 13K & action recognition & FVD & Manual caption annotation. & Strong background bias misleads action recognition.\\

        ActivityNet~\cite{caba2015activitynet}~\href{http://activity-net.org/}{[Link]} & 2015 & 200 & - & 28K & action recognition & FVD & Manual caption annotation. & Background bias.\\

        MSR-VTT~\cite{xu2016msr}~\href{https://www.kaggle.com/datasets/vishnutheepb/msrvtt?resource=download}{[Link]} &2016 & 20 & - & 10K & video-language dataset & FVD,  CLIP-Score & Manual caption annotation. Resolution: 240p. & Noisy and duplicate captions.\\
        HowTo100M~\cite{miech2019howto100m}~\href{https://www.di.ens.fr/willow/research/howto100m/}{[Link]} &2019 &  12 & 154 & 136M & video-language dataset & FVD,  CLIP-Score & ASR annotation. Resolution: 240p. & Noisy and misaligned narration.\\

        WebVideo-10M~\cite{bain2021frozen}~\href{https://github.com/m-bain/webvid} &2021 & - & - & 10M & video-language dataset & FVD,  CLIP-Score & Alt-text caption annotation. Resolution: 360p. & Low-resolution and watermarked videos. \\

        HD-VILA-100M~\cite{xue2022hdvila}~\href{https://github.com/microsoft/XPretrain/blob/main/hd-vila-100m/README.md}{[Link]} &2022 & 15 & - & 103M & video-language dataset & FVD,  CLIP-Score & ASR annotation. Resolution: 720p. & Noisy and misaligned ASR subtitles.\\

        InternVid~\cite{wang2023internvid}~\href{https://github.com/OpenGVLab/InternVideo/tree/main/Data/InternVid}{[Link]} &2023 & 16 & - & 7M & video-language dataset & FVD,  CLIP-Score & Automatic caption annotation. & Limited diversity.\\

        Panda-70M~\cite{chen2024panda}~\href{https://snap-research.github.io/Panda-70M/}{[Link]} &2024 & - & - & 70M & video-language dataset & FVD,  CLIP-Score & Automatic caption annotation. Resolution: 720p. & Limited diversity.\\

        Koala-36M~\cite{wang2024koala}~\href{https://huggingface.co/datasets/Koala-36M/Koala-36M-v1}{[Link]} &2024 & - & - & 36M & video-language dataset & FVD,  CLIP-Score & Automatic and manual caption annotation. Resolution: 720p. & Inherent biases from its source corpus. 
        \\

        \midrule

        \belowrulesepcolor{gray!15!}
        \multicolumn{9}{c}{\textbf{3D Generation (Sec.~\ref{sec:3d})}} \\ \aboverulesepcolor{gray!15!} \midrule
        DeepFashion~\cite{liu2016deepfashion}~\href{https://mmlab.ie.cuhk.edu.hk/projects/DeepFashion.html}{[Link]} &2016 & 50 & 300K & 800K & single-view images  & -  & Human clothing dataset. & No 3D data.  \\

        SHHQ~\cite{fu2022stylegan}~\href{https://stylegan-human.github.io/data.html}{[Link]} &2022 & - & - & 230K & single-view images  & - & Human body dataset. & (1) No 3D data. (2) Inherent biases from FFHQ. \\
        
        CO3D~\cite{reizenstein2021common}~\href{https://ai.meta.com/datasets/co3d-downloads/}{[Link]} &2021 & 50 & 19K & 1.5M & multi-view images  & PSNR/LPIPS; IoU, etc.  &  Annotated with camera poses and point clouds. & Lack high-fidelity geometry/textures.\\
        RTMV~\cite{tremblay2022rtmv}~\href{https://www.cs.umd.edu/~mmeshry/projects/rtmv/}{[Link]} &2021 & - & 2K & 300K &  multi-view images  & PSNR/LPIPS; MAE, etc.  & Large-scale synthetic dataset with high-resolution images. & Synthetic data cannot fully capture real-world complexity.\\
        MVImgNet~\cite{yu2023mvimgnet}~\href{https://gaplab.cuhk.edu.cn/projects/MVImgNet/}{[Link]} &2023 & 238 & 219K & 6.5M & multi-view images  & PSNR/LPIPS & Annotated with camera poses and point clouds.  & Lack high-fidelity geometry/textures.\\
        
        uCO3D~\cite{liu24uco3d}~\href{https://github.com/facebookresearch/uco3d}{[Link]} &2025 & 1K & 170K & 170K & multi-view images  & PSNR/LPIPS; IoU & Annotated with camera poses, point clouds, depth maps, caption, and 3DGS. & Lack high-fidelity geometry/textures. \\
        
        ShapeNet~\cite{chang2015shapenet}~\href{https://shapenet.org/}{[Link]} &2015 & 3K & 51K & & 3D data  & - & Large-scale synthetic CAD dataset. &  Synthetic data cannot fully capture real-world complexity. \\

        3D-Future~\cite{fu20213d}~\href{https://tianchi.aliyun.com/specials/promotion/alibaba-3d-future}{[Link]} &2020 & 34 & 10K & - & 3D data  & CLIP Similarity & High-quality 3D models with textures and attributes. & (1) Lack object diversity. (2) Synthetic data cannot fully capture real-world complexity.\\
        GSO~\cite{downs2022google}~\href{https://app.gazebosim.org/GoogleResearch/fuel/collections/}{[Link]} &2022 & 17 & 1K & - & 3D data  & F-Score/ CLIP Similarity & High-fidelity geometry/textures & Limited to household objects. \\
        
        Objaverse~\cite{deitke2023objaverse}~\href{https://objaverse.allenai.org/objaverse-1.0}{[Link]} &2022 & 21K & 818K & - & 3D data  & F-Score/ CLIP Similarity & Large-scale dataset. & Variable data quality. \\

        Objaverse-XL~\cite{deitke2024objaverse}~\href{https://objaverse.allenai.org/docs/objaverse-xl}{[Link]} &2023 & - & 10.2M & - & 3D data  & F-Score/ CLIP Similarity & Large scale and high diversity. & Variable data quality. \\
        Cap3D~\cite{luo2024scalable}~\href{https://huggingface.co/datasets/tiange/Cap3D}{[Link]} &2023 & - & 785k & - & 3D-text pairs  & FID/CLIP etc. & Rich text annotations & Inherent bias in captions from pretrained models.  \\
        
        Animal2400~\cite{lorraine2023att3d} &2023 & - & 2400 & - & prompts  & CLIP R-Precision & 10 animals, 8 activities, 6 themes, 5 hats. & Limited object diversity. \\
        
        DF27/DF415~\cite{poole2024dreamfusion}~\href{https://dreamfusion3d.github.io/}{[Link]} &2024 & - & 27/415 &- & prompts  & CLIP R-Precision &  - & Limited to synthetic or artist-created objects. \\

        \midrule

        \belowrulesepcolor{gray!15!}
        \multicolumn{9}{c}{\textbf{4D Generation (Sec.~\ref{sec:4d})}} \\ \aboverulesepcolor{gray!15!} \midrule
        Consistent4D~\cite{jiang2023consistent4d}~\href{https://consistent4d.github.io/}{[Link]} &2023 & - & - & 12(in-the-wild), 14(synthetic) & multi-view videos  & LPIPS/CLIP & \textit{Video-to-4D Generation.} & - \\
        
        Diffusion4D~\cite{liang2024diffusion4d}~\href{https://huggingface.co/datasets/hw-liang/Diffusion4D}{[Link]} &2024 & - & 365K & 365K & dynamic 3D data  & - & Collected from Objaverse-1.0 (42K) and Objaverse-xl (323K) & - \\

        MV-Video~\cite{jiang2024animate3d}~\href{https://huggingface.co/datasets/yanqinJiang/MV-Video}{[Link]} &2024 & - & 53K & 53K & dynamic 3D data  & - &  Annotated by MiniGPT4-Video~\cite{ataallah2024minigpt4} & Inaccurate captions.\\

        CamVid-30K~\cite{zhao2024genxd}~\href{https://huggingface.co/datasets/Yuyang-z/CamVid-30K}{[Link]} &2024 & - & - & 30K & 4D data  & - &  Real-world 4D scene dataset & - \\

        4D-DRESS~\cite{wang20244d}~\href{https://4d-dress.ait.ethz.ch/}{[Link]} &2024 & - & - & 64(outfits) & human clothing 4D data  & CD & Total number of 3D human frames: 78k. & Manual-heavy pipeline limits scalability. \\

        \bottomrule[1pt]
    \end{tabular}}

      \begin{tablenotes}
        \footnotesize
        \item "ASR" stands for automatic speech recognition, "pcl" for point clouds, and "CD" for Chamfer Distance.
      \end{tablenotes}
    \end{threeparttable}
    }
    \label{tab:dataset}
\end{table*}
\begin{table*}[!t]
\centering\small
\caption{Summary of common evaluation metrics.}
\setlength{\tabcolsep}{2pt}
\scalebox{0.68}{
\begin{tabular}{llp{1.25\textwidth}}
\toprule[1pt]
& \textbf{Notation} & \textbf{Description} \\ \midrule

\belowrulesepcolor{gray!15!}
\multicolumn{3}{c}{\textbf{Quality (quantitative analysis)}} \\ \aboverulesepcolor{gray!15!} \midrule

\multirow{5}{*}{\rotatebox{+90}{\hspace{-0cm}\textbf{Image-level}}}
& PSNR $\uparrow$ & \textit{Image Fidelity/Similarity.} \textbf{P}eak \textbf{S}ignal-to-\textbf{N}oise \textbf{R}atio: the ratio between the peak signal and the Mean Squared
Error (MSE).\\ 
& SSIM $\uparrow$ & \textit{Image Fidelity/Similarity.} \textbf{S}tructural \textbf{S}imilarity \textbf{I}ndex \textbf{M}easure~\cite{wang2004image}: evaluating brightness, contrast, and structural features between generated and original images.\\ 
& LPIPS $\downarrow$ & \textit{Image Fidelity/Similarity.} \textbf{L}earned \textbf{P}erceptual \textbf{I}mage \textbf{P}atch \textbf{S}imilarity~\cite{zhang2018perceptual}: metric computed with a model trained on labeled human-judged perceptual similarity.\\  
& FID $\downarrow$ & \textit{Image Fidelity/Similarity.} \textbf{F}réchet \textbf{I}nception \textbf{D}istance~\cite{heusel2017gans}: comparison of generated and GT image distributions.\\ 
& IS $\uparrow$ & \textit{Image Fidelity/Similarity.} \textbf{I}nception \textbf{S}core~\cite{salimans2016improved}: metric only evaluated image distributions and calculated by pretrained Inception v3 model. \vspace{0.1em}\\  \midrule

\multirow{5}{*}{\rotatebox{+90}{\hspace{-0cm}\textbf{Video-level}}}
& FVD $\downarrow$ & \textit{Video Fidelity.} \textbf{F}réchet \textbf{V}ideo \textbf{D}istance~\cite{Unterthiner2019FVDAN}:  the Inflated-3D Convnets (I3D) pretrained model to compute their means and covariance matrices for scores.\\ 
& KVD $\downarrow$  & \textit{Video Fidelity.} \textbf{K}ernel \textbf{V}ideo \textbf{D}istance~\cite{Unterthiner2019FVDAN}: an alternative to FVD proposed in the same work, using a polynomial kernel.\\ 
& Video IS $\uparrow$ & \textit{Video Fidelity.} \textbf{Video} \textbf{I}nception \textbf{S}core~\cite{saito2020train}: the
inception score of videos with the features extracted from C3D~\cite{tran2015learning}. \\ 
& FVMD $\downarrow$  & \textit{Video Fidelity.} \textbf{F}réchet \textbf{V}ideo \textbf{M}otion \textbf{D}istance~\cite{liu2024fr}: metric focused on \textit{temporal consistency}, measuring the similarity between motion features of generated and reference videos using Fréchet Distance. \\  \midrule

\belowrulesepcolor{gray!15!}
\multicolumn{3}{c}{\textbf{Alignment (quantitative analysis)}} \\ \aboverulesepcolor{gray!15!} \midrule

\multirow{7}{*}{\rotatebox{+90}{\hspace{-0cm}\textbf{Semantics}}}
& CLIP Similarity~\cite{hessel2021clipscore} $\uparrow$ & \textit{Text-Image Alignment.} Same as CLIP-Score, defined as a text-to-image similarity metric, measuring the (cosine) similarity of the embedded image and text prompt.\\ 
& CLIP R-Precision~\cite{park2021benchmark} $\uparrow$ &  \textit{Text-Image Alignment.} The CLIP model’s accuracy at classifying the correct text input of a rendered image from amongst a set of distractor query prompts. \\ 
& X-CLIP~\cite{ni2022expanding} $\uparrow$ &  \textit{Text-Image Alignment.} Measured by a video-based CLIP model finetuned on text-video data. \\ 
& CLIP-T $\uparrow$ &  \textit{CLIP Textual Alignment.} The average cosine similarity
between the generated frames and text prompt with CLIP ViT-B/32~\cite{dosovitskiy2020image} image and text models.\\ 
& CLIP-I $\uparrow$ &  \textit{CLIP Image Alignment.} The average cosine similarity between the generated frames and subject images with CLIP ViT-B/32 image model.\\ 
& DINO-I $\uparrow$ &  \textit{DINO~\cite{caron2021emerging} Image Alignment.} The average visual similarity between generated frames and reference images with DINO ViT-S/16 model.\\ 
\midrule

\belowrulesepcolor{gray!15!}
\multicolumn{3}{c}{\textbf{User study (qualitative analysis)}} \\ \aboverulesepcolor{gray!15!} \midrule
& AQ $\uparrow$ & \textit{Human Preference.} \textbf{A}ppearance \textbf{Q}uality: percentage unit ($\%$).\\ 
& GQ $\uparrow$ & \textit{Human Preference.} \textbf{G}eometry \textbf{Q}uality: percentage unit ($\%$).\\
& DQ $\uparrow$ & \textit{Human Preference.} \textbf{D}ynamics \textbf{Q}uality: percentage unit ($\%$).\\
& TA $\uparrow$ & \textit{Human Preference.} \textbf{T}ext \textbf{A}lignment: percentage unit ($\%$).\\

\bottomrule
\end{tabular}
}
\label{tab:metrics}
\end{table*}

\label{sec:datasets}
In this section, we summarize the commonly used datasets for 2D, video, 3D, and 4D generation in Table~\ref{tab:dataset}. Then, we present a unified and comprehensive summary of evaluation metrics in Table~\ref{tab:metrics}. For quantitative analysis, we evaluate metrics from two perspectives: 1) \textit{Quality}: assessing the perceptual quality of the synthesized data, independent of input conditions (e.g., text prompts). 2) \textit{Alignment}: measuring condition consistency, i.e., how well the generated data matches the user's intended input. For qualitative analysis, the visual quality of generated results plays a critical role in assessing methods. Therefore, we include some human preference-based indicators as references for more effectively conducting user studies, enabling more convincing qualitative analysis results. Besides, we advocate emphasizing the practical challenges in deploying generative models, especially those related to computational efficiency. Many state-of-the-art methods require extensive GPU resources and long inference times, which hinder their accessibility and scalability in real-world applications. These factors are not always reflected in evaluation metrics but are essential when considering deployment in resource-constrained environments or interactive systems. We also encourage future benchmarks to include runtime, memory usage, and training cost to better reflect the practical feasibility of generative models.

\tikzset{
    my-box/.style={
        rectangle,
        draw=hidden-draw,
        rounded corners,
        text opacity=1,
        minimum height=1.5em,
        minimum width=5em,
        inner sep=2pt,
        align=center,
        fill opacity=.5,
        line width=0.8pt
    },
    leaf-video/.style={
        my-box,
        minimum height=1.5em,
        fill=hidden-video!80,
        text=black,
        align=left,
        font=\tiny,
        inner xsep=2pt,
        inner ysep=4pt,
        line width=0.8pt
    },
    leaf/.style={
        my-box,
        fill=white, 
        text=black
    }
}

\begin{figure*}[!t]
    \centering
    \begin{adjustbox}{width=0.95\textwidth}
        \begin{forest}
            forked edges,
            for tree={
                grow=east,
                reversed=true,
                anchor=base west,
                parent anchor=east,
                child anchor=west,
                base=left,
                font=\tiny,
                rectangle,
                draw=hidden-draw,
                rounded corners,
                align=left,
                minimum width=2em,
                edge+={gray, line width=0.8pt},
                s sep=3pt,
                inner xsep=2pt,
                inner ysep=3pt,
                line width=0.8pt,
                ver/.style={rotate=90, child anchor=north, parent anchor=south, anchor=center},
            },
            [\textbf{Approaches}, ver
                [\textbf{(1) VAE- and GAN-based}, leaf, text width=6.5em
                    [
                         {SV2P~\cite{babaeizadeh2017stochastic}, FitVid~\cite{babaeizadeh2021fitvid}, MoCoGAN~\cite{mocogan}, StyleGAN-V~\cite{skorokhodov2022stylegan}, DIGAN~\cite{yu2022generating}, StyleInV~\cite{wang2023styleinv}}\\
                         {\textit{Remarks}: Limited resolution and domain.}
                            , leaf-video, text width=22.5em
                    ]
                ]
                [\textbf{(2) Diffusion-based}, leaf, text width=5em
                    [    {\textbf{(a) U-Net-based Models}: 
                         VDM~\cite{ho2022video}, Make-A-Video~\cite{singer2022make}, Imagen Video~\cite{ho2022imagen}, MagicVideo~\cite{zhou2022magicvideo}, GEN-1~\cite{esser2023structure}, PYoCo~\cite{ge2023preserve}, \\ Align-your-Latents~\cite{blattmann2023align}, VideoComposer~\cite{wang2024videocomposer}, AnimateDiff~\cite{guo2023animatediff}, PixelDance~\cite{zeng2024make}, Emu Video~\cite{girdhar2023emu}, Lumiere~\cite{bar2024lumiere}\\
                         \textit{Remarks}: Image pretrained models, per-frame VAE.}
                            , leaf-video, text width=29em
                    ] 
                    [    {\textbf{(b) Transformer-based Models}: 
                       VDT~\cite{lu2023vdt}, W.A.L.T~\cite{gupta2024photorealistic}, GenTron~\cite{chen2024gentron}, Luminia-T2X~\cite{gao2024lumina}, \\
                       CogVideoX~\cite{yang2024cogvideox}, Sora~\cite{videoworldsimulators2024}, Cosmos WFM~\cite{agarwal2025cosmos}\\
                         \textit{Remarks}: Spatiotemporal attention.}
                            , leaf-video, text width=22em
                    ] 
                ]
                [\textbf{(3) Autoregressive-based}, leaf, text width=6.5em
                    [
                         {MAGVIT~\cite{yu2023magvit}, CogVideo~\cite{hong2022cogvideolargescalepretrainingtexttovideo}, VideoPoet~\cite{kondratyuk2023videopoet},
                         Cosmos WFM~\cite{agarwal2025cosmos}
                         }\\
                          {\textit{Remarks}: Decoder-only LLM architecture.}
                            , leaf-video, text width=16em
                    ]
                ]
            ]
        \end{forest}
    \end{adjustbox}
    \caption{Overview of text-to-video generation technologies categorized by three main approaches.}

    \label{fig:text-to-video}
\end{figure*}

\begin{table*}[t]
    \caption{Recent text-to-3D, image-to-3D and video-to-3D generation methods. 
    }
    \small
    \center
    \setlength{\tabcolsep}{5pt}
    \scalebox{0.85}{
    \begin{tabular}{c|c|r|c|c|c|c}
    \hline
    \multicolumn{2}{c|}{\textbf{Paradigms}} & \textbf{Methods} & \textbf{Paper} & \textbf{Code} & \textbf{Representations} & \textbf{Objectives} \\
    \hline
    \multirow{25}{*}{\rotatebox{60}{\textbf{Text-to-3D}}} 
    & \multirow{10}{*}{\rotatebox{60}{Feedforward}} & {Point$\cdot$E} ~\cite{nichol2022point} & \href{https://arxiv.org/abs/2212.08751} {\color{green}{Link}} & \href{https://github.com/openai/point-e} {\color{green}{Link}} & Point Clouds & MAE \\
    & & 3D-LDM ~\cite{nam20223d} & \href{https://arxiv.org/abs/2212.00842}{\color{green}{Link}} & - & SDF & reconstruction+regularization\\
    & & ATT3D~\cite{lorraine2023att3d} & \href{https://arxiv.org/abs/2306.07349}{\color{green}{Link}}& \href{https://research.nvidia.com/labs/toronto-ai/ATT3D/}{\color{green}{Link}} & Instant-NGP model & SDS \\
    & & Diffusion-SDF ~\cite{li2023diffusion} & \href{https://arxiv.org/abs/2212.03293}{\color{green}{Link}} & \href{https://github.com/ttlmh/Diffusion-SDF}{\color{green}{Link}} & SDF & reconstruction+KL-Divergence\\
    & & LATTE3D~\cite{xie2024latte3d} & \href{https://arxiv.org/abs/2403.15385}{\color{green}{Link}}&\href{https://research.nvidia.com/labs/toronto-ai/LATTE3D/}{\color{green}{Link}}& NeRF/SDF & SDS+regularization \\
    & & MeshDiffusion ~\cite{liu2023meshdiffusion} & \href{https://arxiv.org/abs/2303.08133}{\color{green}{Link}} & \href{https://github.com/lzzcd001/MeshDiffusion/}{\color{green}{Link}} & Mesh & diffusion loss\\
    & & {Shap$\cdot$E} ~\cite{shape-e} & \href{https://arxiv.org/abs/2305.02463}{\color{green}{Link}} & \href{https://github.com/openai/shap-e}{\color{green}{Link}} & NeRF & rendering loss\\
    & & Hyperfields~\cite{babu2024hyperfields} & \href{https://arxiv.org/abs/2310.17075}{\color{green}{Link}} & \href{https://github.com/threedle/hyperfields}{\color{green}{Link}} & NeRF  & SDS\\
    & & Michelangelo~\cite{zhao2023michelangelo} & \href{https://arxiv.org/abs/2306.17115}{\color{green}{Link}} & \href{https://github.com/NeuralCarver/Michelangelo}{\color{green}{Link}} & Occupancy Field & text-image-shape contrastive + BCE\\
    & & Atom~\cite{qian2024atom} & \href{https://arxiv.org/abs/2402.00867}{\color{green}{Link}}& - & Triplane & SDS \\
    \cdashline{2-7} 
    & \multirow{12}{*}{\rotatebox{60}{Optimization}} 
    & Magic3D~\cite{lin2023magic3d} & \href{https://arxiv.org/abs/2211.10440}{Link} & \href{https://research.nvidia.com/labs/dir/magic3d/}{Link} & NeRF/DMTet & SDS \\
    & & Dream3D~\cite{xu2023dream3d} & \href{https://arxiv.org/abs/2212.14704}{Link} & - & NeRF & reconstruction+regularization \\
    & & Fantasia3D~\cite{chen2023fantasia3d} & \href{https://arxiv.org/abs/2303.13873}{Link} & \href{https://github.com/Gorilla-Lab-SCUT/Fantasia3D}{Link} & DMTet & SDS \\
    & & DreamFusion~\cite{poole2024dreamfusion} & \href{https://arxiv.org/abs/2209.14988}{Link} & - & NeRF & SDS \\
    & & ProlificDreamer~\cite{wang2024prolificdreamer} & \href{https://arxiv.org/abs/2305.16213}{Link} & \href{https://github.com/thu-ml/prolificdreamer}{Link} & NeRF & VSD \\
    & & PI3D~\cite{liu2024pi3d} & \href{https://arxiv.org/abs/2312.09069}{Link} & - & Triplane & SDS \\
    & & VP3D~\cite{chen2024vp3d} & \href{https://arxiv.org/abs/2403.17001}{Link} & - & NeRF & SDS \\
    & & MVDream~\cite{shi2023mvdream} & \href{https://arxiv.org/abs/2308.16512}{Link} &\href{https://mv-dream.github.io/}{Link} & NeRF & SDS \\
    & & DreamGuassian~\cite{tang2024dreamgaussian} & \href{https://arxiv.org/abs/2309.16653}{Link} & \href{https://github.com/dreamgaussian/dreamgaussian}{Link} & 3DGS & SDS \\
    & & GaussianDreamer~\cite{yi2023gaussiandreamer} & \href{https://arxiv.org/abs/2310.08529}{Link} & \href{https://github.com/hustvl/GaussianDreamer}{Link} & 3DGS & SDS \\
    & & GSGEN~\cite{chen2024text} & \href{https://arxiv.org/abs/2309.16585}{Link} & \href{https://github.com/gsgen3d/gsgen}{Link} & 3DGS & SDS \\
    & & Sculpt3D~\cite{chen2024sculpt3d} & \href{https://arxiv.org/abs/2403.09140}{Link} & \href{https://github.com/StellarCheng/Scuplt_3d/tree/main}{Link} & NeRF & diffusion loss \\
    \cdashline{2-7} 
    & \multirow{3}{*}{\rotatebox{60}{MVS}} & Instant3D~\cite{li2024instant3d} & \href{https://arxiv.org/abs/2311.06214} {\color{blue}{Link}} & \href{https://jiahao.ai/instant3d/} {\color{blue}{Link}} & Triplane & MSE+LPIPS \\
    & & Direct2.5~\cite{lu2024direct2} & \href{https://arxiv.org/abs/2311.15980} {\color{blue}{Link}} & \href{https://github.com/apple/ml-direct2.5} {\color{blue}{Link}} & Multi-view normal maps & normal rendering+alpha mask loss \\
    & & Sherpa3D~\cite{liu2024sherpa3d} & \href{https://arxiv.org/abs/2312.06655} {\color{blue}{Link}} & \href{https://github.com/liuff19/Sherpa3D} {\color{blue}{Link}} & DMTet & SDS \\
    \hline
    \multirow{20}{*}{\rotatebox{60}{\textbf{Image-to-3D}}} 
    & \multirow{7}{*}{\rotatebox{60}{Feedforward}} & 3DGen~\cite{gupta20233dgen} & \href{https://arxiv.org/abs/2303.05371} {\color{green}{Link}} & - & DMTet & rendering loss\\
    & & {Shap$\cdot$E}~\cite{shape-e} & \href{https://arxiv.org/abs/2305.02463}{\color{green}{Link}} & \href{https://github.com/openai/shap-e}{\color{green}{Link}} & NeRF & rendering loss\\
    & & Michelangelo~\cite{zhao2023michelangelo} & \href{https://arxiv.org/abs/2306.17115}{\color{green}{Link}} & \href{https://github.com/NeuralCarver/Michelangelo}{\color{green}{Link}} & Occupancy Field & text-image-shape contrastive + BCE\\
    & & Clay~\cite{zhang2024clay} & \href{https://arxiv.org/abs/2406.13897}{\color{green}{Link}} & \href{https://sites.google.com/view/clay-3dlm}{\color{green}{Link}} & Occupancy Field & BCE\\
    & & CraftsMan~\cite{li2024craftsman} & \href{https://arxiv.org/abs/2405.14979}{\color{green}{Link}} & \href{https://craftsman3d.github.io/}{\color{green}{Link}} & Occupancy Field & BCE\\
    & & Direct3D~\cite{shuang2025direct3d} & \href{https://arxiv.org/abs/2405.14832}{\color{green}{Link}} & \href{https://nju-3dv.github.io/projects/Direct3D/}{\color{green}{Link}} & Occupancy Field & BCE\\
    & &Trellis~\cite{xiang2025structured} & \href{https://arxiv.org/abs/2412.01506}{\color{green}{Link}} & \href{https://trellis3d.github.io/}{\color{green}{Link}} & 3DGS/RF/Mesh & MSE\\
    \cdashline{2-7}
    & \multirow{9}{*}{\rotatebox{60}{Optimization}} & RealFusion~\cite{melas2023realfusion} & \href{https://arxiv.org/abs/2302.10663}{Link} &\href{https://github.com/lukemelas/realfusion}{Link} & NeRF & SDS/Image Reconstruction \\
    & & Zero123~\cite{liu2023zero} & \href{https://arxiv.org/abs/2303.11328}{Link} & \href{https://zero123.cs.columbia.edu/}{Link} & NeRF & SDS\\
    & & Magic123~\cite{qian2023magic123} & \href{https://arxiv.org/abs/2306.17843}{Link} & \href{https://guochengqian.github.io/project/magic123/}{Link} & NeRF & SDS\\
    & & Syncdreamer~\cite{liu2024syncdreamer} & \href{https://arxiv.org/abs/2309.03453}{Link} & \href{https://liuyuan-pal.github.io/SyncDreamer/}{Link} & NeRF & SDS\\
    & & Consistent123~\cite{weng2023consistent123} & \href{https://arxiv.org/abs/2310.08092}{Link} & \href{https://consistent-123.github.io/}{Link} & NeRF & SDS\\
    & & Toss~\cite{shi2023toss} & \href{https://arxiv.org/abs/2310.10644}{Link} & \href{https://toss3d.github.io/}{Link} & NeRF & SDS\\
    & & ImageDream~\cite{wang2023imagedream} & \href{https://arxiv.org/abs/2312.02201}{Link} & \href{https://image-dream.github.io/}{Link} & NeRF & SDS\\
    & & IPDreamer~\cite{zeng2023ipdreamer} & \href{https://arxiv.org/abs/2310.05375}{Link} & \href{https://github.com/zengbohan0217/IPDreamer}{Link} & NeRF & SDS \\
    & & Wonder3D~\cite{long2024wonder3d} & \href{https://arxiv.org/abs/2310.15008}{Link} & \href{https://www.xxlong.site/Wonder3D/}{Link} & Mesh & Image/Normal MSE\\
    \cdashline{2-7} 
    & \multirow{5}{*}{\rotatebox{60}{MVS}} & One-2-3-45~\cite{liu2024one} & \href{https://arxiv.org/abs/2306.16928} {\color{blue}{Link}} & \href{https://one-2-3-45.github.io/} {\color{blue}{Link}} & SDF & Image/Depth MSE \\
    & & CRM~\cite{wang2024crm} & \href{https://arxiv.org/abs/2403.05034} {\color{blue}{Link}} & \href{https://ml.cs.tsinghua.edu.cn/~zhengyi/CRM/} {\color{blue}{Link}} & Triplane & Image/Depth MSE \\
    & & InstantMesh~\cite{xu2024instantmesh} & \href{https://arxiv.org/abs/2404.07191} {\color{blue}{Link}} & \href{https://github.com/TencentARC/InstantMesh} {\color{blue}{Link}} & Triplane & Image/Depth/Normal MSE \\
    & & LRM~\cite{hong2023lrm} & \href{https://arxiv.org/abs/2311.04400} {\color{blue}{Link}} & \href{https://yiconghong.me/LRM/} {\color{blue}{Link}} & NeRF & MSE \\
    & & Unique3D~\cite{wu2024unique3d} & \href{https://arxiv.org/abs/2405.20343} {\color{blue}{Link}} & \href{https://wukailu.github.io/Unique3D/} {\color{blue}{Link}} & Mesh & Image/Normal MSE \\
    \hline
    \multirow{5}{*}{\rotatebox{60}{\textbf{Video-to-3D}}} 
    & \multirow{5}{*}{\rotatebox{60}{MVS}} & ViVid-1-to-3~\cite{kwak2024vivid} & \href{https://arxiv.org/abs/2312.01305} {\color{blue}{Link}} & \href{https://ubc-vision.github.io/vivid123/} {\color{blue}{Link}} & Multi-view images & - \\
    & & IM-3D~\cite{melas20243d} & \href{https://arxiv.org/abs/2402.08682} {\color{blue}{Link}} & \href{https://lukemelas.github.io/IM-3D/} {\color{blue}{Link}} & 3DGS & MSE+LPIPS \\
    & & V3D~\cite{chen2024v3d} & \href{https://arxiv.org/abs/2403.06738} {\color{blue}{Link}} & \href{https://heheyas.github.io/V3D/} {\color{blue}{Link}} & Mesh & MSE+LPIPS \\
    & & SV3D~\cite{voleti2025sv3d} & \href{https://arxiv.org/abs/2403.12008} {\color{blue}{Link}} & \href{https://sv3d.github.io/} {\color{blue}{Link}} & NeRF / DMTet & MSE+LPIPS \\
    & & CAT3D~\cite{gao2024cat3d} & \href{https://arxiv.org/abs/2405.10314} {\color{blue}{Link}} & \href{https://cat3d.github.io/} {\color{blue}{Link}} & NeRF & MSE+LPIPS \\
    
    \hline
    \end{tabular}
    }
    \label{tab:text-to-3d}
\end{table*}
\begin{table*}[t]
    \caption{Representative works of 4D generation methods. "Rep" stands for representations.}
    \small
    \center
    \vspace{-1em}
    \setlength{\tabcolsep}{5pt}
    \scalebox{0.7}{
    \begin{threeparttable}
    \begin{tabular}{c|r|c|c|c|c|c}
    \hline
    \textbf{Approaches} & \textbf{Methods} & \textbf{Paper} & \textbf{Code} &\textbf{Representations} & \textbf{Priors / Models} & \textbf{Objectives} \\
    \hline
    \multirow{4}{*}{\rotatebox{60}{Feedforward}} & Control4D~\cite{shao2024control4d} & \href{https://arxiv.org/abs/2305.20082} {\color{green}{Link}} & - & GaussianPlanes & GAN-based & GAN objective \\
    & Animate3D~\cite{jiang2024animate3d} & \href{https://arxiv.org/abs/2407.11398}{\color{green}{Link}} & \href{https://github.com/yanqinJiang/Animate3D}{\color{green}{Link}} & 4DGS & Diffusion-based & latent diffusion loss \\
    & Vidu4D~\cite{wang2024vidu4d} & \href{https://arxiv.org/abs/2405.16822}{\color{green}{Link}} & \href{https://github.com/yikaiw/vidu4d}{\color{green}{Link}} & Dynamic Gaussian Surfels & Diffusion-based & reconstruction/regularization loss \\
    & Diffusion4D~\cite{liang2024diffusion4d} & \href{https://arxiv.org/abs/2405.16645}{\color{green}{Link}} & \href{https://github.com/VITA-Group/Diffusion4D}{\color{green}{Link}} & 4DGS & Diffusion-based & latent diffusion + motion reconstruction loss  \\
    & L4GM~\cite{ren2024l4gm} & \href{https://arxiv.org/abs/2406.10324}{\color{green}{Link}} & \href{https://github.com/nv-tlabs/L4GM-official}{\color{green}{Link}} & 3DGS & VAE-based  & MSE+LPIPS \\
    & GenXD~\cite{zhao2024genxd} & \href{https://arxiv.org/abs/2411.02319}{\color{green}{Link}} & \href{https://github.com/HeliosZhao/GenXD}{\color{green}{Link}} & 4DGS & Diffusion-based  & SSIM+LPIPS \\
    & CAT4D~\cite{wu2025cat4d} & \href{https://arxiv.org/abs/2411.18613}{\color{green}{Link}} & - & Deformable 3DGS & Diffusion-based  & LPIPS \\
    & 4Real-Video~\cite{wang20254real} & \href{https://arxiv.org/abs/2412.04462}{\color{green}{Link}} & - & Deformable 3DGS & Diffusion-based  & velocity matching loss of rectified flow~\cite{liu2022flow} \\
    
    \cdashline{1-7} 
    
    \multirow{13}{*}{\rotatebox{60}{Optimization}} & MAV3D~\cite{singer2023text} & \href{https://arxiv.org/abs/2301.11280}{Link} & - & HexPlane-/NeRF-based 4D Rep & T2I/T2V & SDS \\
    & 4D-fy~\cite{bah20244dfy} & \href{https://arxiv.org/abs/2311.17984}{Link} & \href{https://github.com/sherwinbahmani/4dfy}{Link} & Hash-/NeRF-based 4D Rep & MVDream/T2I/T2V & Hybrid SDS \\
    & AYG~\cite{ling2024align} & \href{https://arxiv.org/abs/2312.13763}{Link} & - & Deformable 3DGS & MVDream, T2I/T2V & SDS variant(CSD~\cite{yu2023text}) \\
    & Dream-in-4D~\cite{zheng2024unified} & \href{https://arxiv.org/abs/2311.16854}{Link} & \href{https://github.com/NVlabs/dream-in-4d}{Link} & NeRF-based 4D Rep & MVDream/T2I/T2V & Hybrid SDS \\
    & TC4D~\cite{bahmani2025tc4d} & \href{https://arxiv.org/abs/2403.17920}{Link} & \href{https://github.com/sherwinbahmani/tc4d}{Link} & Hash-/NeRF-based 4D Rep & T2V & SDS \\
    & 4Real~\cite{yu20244real} & \href{https://arxiv.org/abs/2406.07472}{Link} & - & Deformable 3DGS & T2V & SDS \\
    & C3V~\cite{zhu2024compositional} & \href{https://arxiv.org/abs/2409.00558}{Link} & - & 3DGS & LDMs~\cite{rombach2022high} & SDS \\
    & Consistent4D~\cite{jiang2023consistent4d} & \href{https://arxiv.org/abs/2311.02848}{Link} & \href{https://github.com/yanqinJiang/Consistent4D}{Link} & D-NeRF & Zero123 & SDS \\
    & SC4D~\cite{wu2025sc4d} &\href{https://arxiv.org/abs/2404.03736}{Link} & \href{https://github.com/JarrentWu1031/SC4D}{Link} & SC-GS~\cite{huang2024sc} & SVD/Zero123 & SDS \\
    & STAG4D~\cite{zeng2025stag4d} & \href{https://arxiv.org/abs/2403.14939}{Link} & \href{https://github.com/zeng-yifei/STAG4D}{Link} & Deformable 3DGS & SVD/Zero123 & SDS \\
    & DreamScene4D~\cite{chu2024dreamscene4d} & \href{https://arxiv.org/abs/2405.02280}{Link} & \href{https://github.com/dreamscene4d/dreamscene4d}{Link} & Deformable 3DGS & SVD/Zero123 & SDS \\
    & 4DM~\cite{zhang20244diffusion} & \href{https://arxiv.org/abs/2405.20674}{Link} & \href{https://github.com/aejion/4Diffusion}{Link} & D-NeRF & SVD & 4D-aware SDS + anchor loss \\
    & DreamMesh4D~\cite{li2024dreammesh4d} & \href{https://arxiv.org/abs/2410.06756}{Link} & \href{https://github.com/WU-CVGL/DreamMesh4D}{Link} & SC-GS+Mesh Rep & Zero123 & SDS  \\
    & 4K4DGen~\cite{li20244k4dgen} & \href{https://arxiv.org/abs/2406.13527}{Link} & \href{https://github.com/ShadowIterator/4K4DGen}{Link} & 4DGS & SVD & L1  \\
    & EG4D~\cite{sun2024eg4d} & \href{https://arxiv.org/abs/2405.18132}{Link} & \href{https://github.com/jasongzy/EG4D}{Link} & 4DGS & SVD & L1  \\
    & AvatarGO~\cite{cao2024avatargo} & \href{https://arxiv.org/abs/2410.07164}{Link} & \href{https://github.com/yukangcao/AvatarGO}{Link} & 3DGS+Hexplane & SVD/Zero123 & SDS  \\
    & Disco4D~\cite{pang2025disco4d} & \href{https://arxiv.org/abs/2409.17280}{Link} & \href{https://github.com/disco-4d/Disco4D}{Link} & 4DGS & SVD & SDS  \\
    & GenMOJO~\cite{chu2025robust} & \href{https://www.arxiv.org/abs/2506.12716}{Link} & \href{https://github.com/genmojo/GenMOJO}{Link} & Deformable 3DGS & SVD/Zero123 & SDS \\
    
    \hline
    \end{tabular}
    \end{threeparttable}
    }
    \label{tab:4d_generation}
\end{table*}

\vfill

\end{document}